%% file: AppliancePlan.tex
\documentclass[sigconf]{acmart}
\AtBeginDocument{%
  }

\copyrightyear{2026}
\acmYear{2026}
\setcopyright{cc}
\setcctype{by}
\acmConference[MM '26]{Proceedings of the 34th ACM International Conference on Multimedia}{November 10--14, 2026}{Rio de Janeiro, Brazil}
\acmBooktitle{Proceedings of the 34th ACM International Conference on Multimedia (MM '26), November 10--14, 2026, Rio de Janeiro, Brazil}
\acmDOI{10.1145/3767308.3836196}
\acmISBN{979-8-4007-2213-4/2026/11}

\usepackage[table]{xcolor}
\usepackage{multirow}
\usepackage{pifont}
\usepackage{xspace}
\newcommand{\dataset}{UseAppliance\xspace}
\newcommand{\model}{AppliancePlan\xspace}
\newcommand{\benchmark}{RealAppliance-Bench\xspace}

\begin{document}

\title[Scaling Manual-Grounded Appliance Manipulation with Data Synthesis and Unified Planning]{Scaling Manual-Grounded Appliance Manipulation with Data Synthesis and Unified Planning}

\author{Yuxing Long}
\authornote{These authors contributed equally to this work.}
\affiliation{%
  \institution{Center on Frontiers of Computing Studies, School of Computer Science, Peking University}
  \city{}
  \country{}}

\author{Lei Kang}
\authornotemark[1]
\affiliation{%
  \institution{Center on Frontiers of Computing Studies, School of Computer Science, Peking University}
  \city{}
  \country{}}

\author{Ziyan Yu}
\affiliation{%
  \institution{Center on Frontiers of Computing Studies, School of Computer Science, Peking University}
  \city{}
  \country{}}

\author{Yuzheng Gao}
\affiliation{%
  \institution{Center on Frontiers of Computing Studies, School of Computer Science, Peking University}
  \city{}
  \country{}}

\author{Bin Cheng}
\affiliation{%
  \institution{Beijing University of Aeronautics and Astronautics}
  \city{}
  \country{}}

\author{Jiyao Zhang}
\affiliation{%
  \institution{Center on Frontiers of Computing Studies, School of Computer Science, Peking University}
  \city{}
  \country{}}

\author{Xiaoqi Li}
\affiliation{%
  \institution{Center on Frontiers of Computing Studies, School of Computer Science, Peking University}
  \city{}
  \country{}}

\author{Haolin Yang}
\affiliation{%
  \institution{Center on Frontiers of Computing Studies, School of Computer Science, Peking University}
  \city{}
  \country{}}

\author{Dongjiang Li}
\affiliation{%
  \institution{Jingdong Technology Information Technology Co., Ltd}
  \city{}
  \country{}}

\author{Hui Shen}
\affiliation{%
  \institution{Jingdong Technology Information Technology Co., Ltd}
  \city{}
  \country{}}

\author{Hao Dong}
\authornote{Corresponding author.}
\affiliation{%
  \institution{Center on Frontiers of Computing Studies, School of Computer Science, Peking University}
  \city{}
  \country{}}
\email{hao.dong@pku.edu.cn}

\renewcommand{\shortauthors}{Long et al.}

\input{sec/0_abstract}


\maketitle

\begin{figure*}[t]
  \centering
  \includegraphics[width=\textwidth]{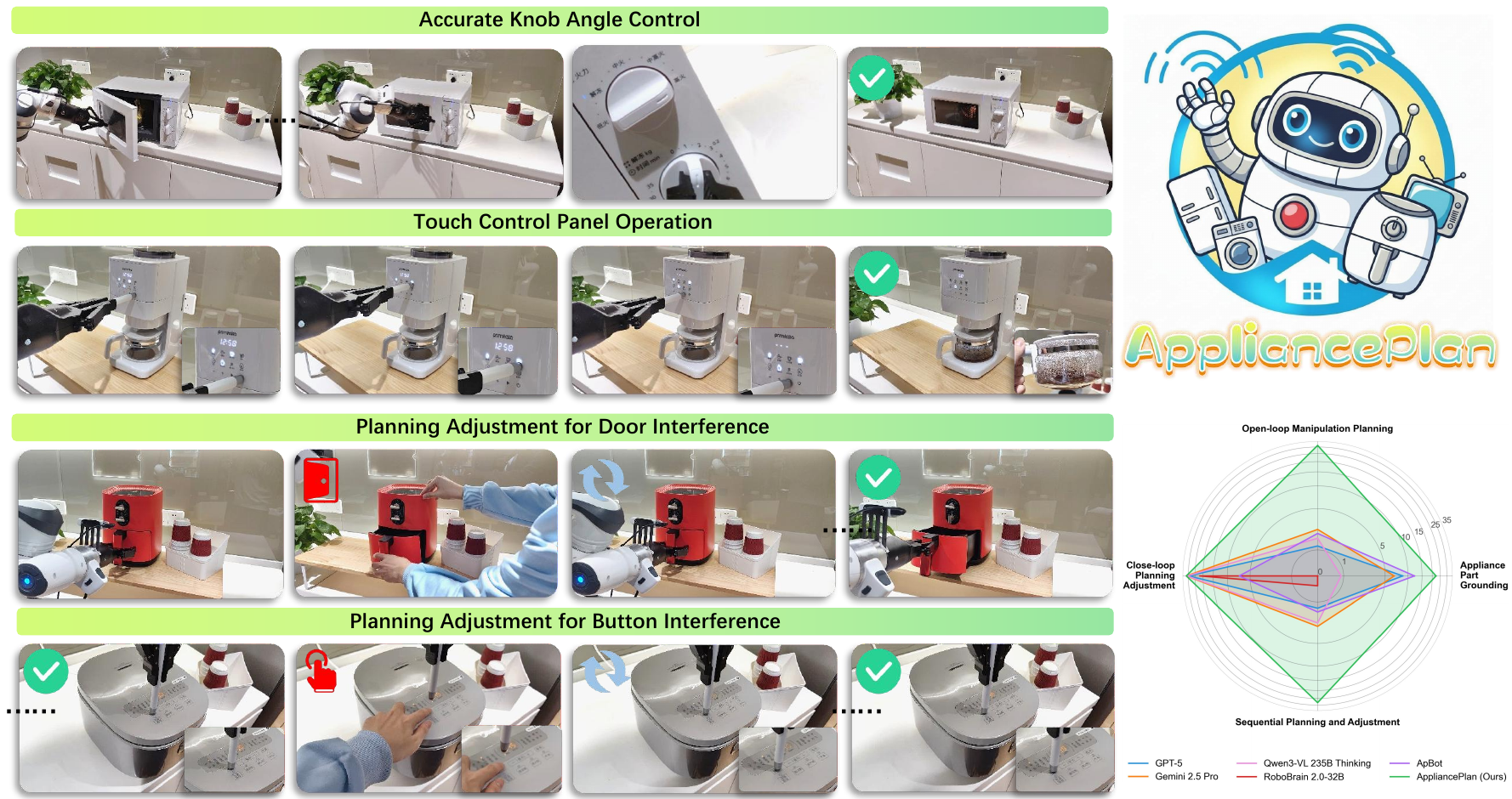}
  \caption{\textbf{Capabilities of our \model model.} Left: \model plans manipulation for diverse appliances and updates the plan when interrupted. Right: \model outperforms strong large-model baselines on \benchmark~\cite{realappliance}.}
  \label{fig:teaser}
\end{figure*}

\input{sec/1_intro}
\input{sec/2_relatedwork}

\input{sec/3_data_pipeline}
\input{sec/4_dataset}

\input{sec/5_model}
\input{sec/6_experiments}

\begin{acks}
This work was supported by the Beijing Natural Science Foundation (L2608141) and the National Natural Science Foundation of China (62136001).
\end{acks}

\bibliographystyle{ACM-Reference-Format}
\bibliography{reference}

\end{document}

%% file: sec/0_abstract.tex
\begin{abstract}
Operating household appliances requires long-horizon planning that is state-dependent and robust to disturbances, yet existing large models fall short, as no sufficiently diverse, task-oriented dataset exists to support such planning. To bridge this gap, we propose MAGE, a scalable data synthesis pipeline that introduces a novel Hierarchical Appliance Graph (HAG) to automatically generate part grounding, long-horizon planning, and closed-loop recovery data from appliance manuals. With MAGE, we build \dataset, the first large-scale dataset for manual-grounded appliance manipulation planning, spanning 22 appliance categories with 89K+ part annotations, 53K+ manipulation tasks, and 33K+ closed-loop adjustment steps. Built on \dataset, we develop \model, an end-to-end model for manual-grounded appliance manipulation planning. On \benchmark, \model with only 7B parameters achieves over 10$\times$ the best baseline on open-loop planning and consistently outperforms state-of-the-art models across all tasks. Real-robot experiments on six household appliances further confirm effective sim-to-real transfer, marking an important step toward general-purpose household robotics.
\end{abstract}

%% file: sec/1_intro.tex
\vspace{-4pt}
\section{Introduction}
\label{sec:intro}
\vspace{-2pt}
Enabling robots to operate household appliances is a key prerequisite for practical home autonomy. Appliance manipulation is substantially harder than generic pick-and-place: appliances vary widely in interface layout, operation logic, and state transitions, and successful execution requires long-horizon, state-dependent procedures that must remain robust to disturbances.

Humans address this by consulting manuals—an approach that naturally decomposes the task into three sub-capabilities: appliance part grounding, open-loop manipulation planning, and closed-loop planning adjustment. \benchmark~\cite{realappliance} provides a unified evaluation protocol for these capabilities, pairing 100 real appliance manuals with high-fidelity digital assets. Yet existing MLLMs and embodied planning models still perform poorly on \benchmark. Prior methods either focus on simple interactions (\emph{e.g.}, pressing a single button)~\cite{li2024manipllm,black2024pi_0,OmniManip} or rely on multi-stage zero-shot pipelines built upon proprietary models, incurring high inference latency and compounding errors~\cite{checkmanual,apbot,OmniManip}. \textbf{The fundamental bottleneck is the lack of large-scale, high-quality training data tailored to manual-grounded appliance manipulation.}

To address this bottleneck, we introduce \textbf{MAGE} (Manual-grounded Appliance data GEneration pipeline), a scalable data synthesis pipeline built on a novel Hierarchical Appliance Graph (HAG). HAG represents each appliance manual as a four-level hierarchy—from document and page structure down to individual parts and state-transition graphs. MAGE collects and classifies manuals via autonomous agents to form the upper layers of each HAG, then completes the lower layers through model-assisted annotation. With the HAG in place, data generation reduces to combinatorial graph traversal—sampling state pairs for plan synthesis and injecting disturbances for closed-loop recovery—with human verification at every stage. This yields three data types: \emph{(i)}~multi-view bounding-box for part grounding; \emph{(ii)}~long-horizon manipulation plans with systematic state-space coverage; and \emph{(iii)}~planning-aligned observation images with closed-loop recovery data.

With MAGE, we build \dataset, the first large-scale dataset for manual-grounded appliance manipulation planning, with four key properties: \emph{(i)~high quality}—Core annotations are human-labeled and double-verified; \emph{(ii)~rich diversity}—spanning 22 appliance categories from manufacturers worldwide with diverse control interfaces and operation procedures; \emph{(iii)~multi-task coverage}—providing unified supervision for all three capabilities with 89K+ part bounding boxes, 53K+ manipulation tasks, and 33K+ closed-loop adjustment steps; and \emph{(iv)~extensibility}—the MAGE pipeline is fully decoupled from specific appliances, allowing new training data to be generated from any appliance manual.

Furthermore, we propose \model, a unified end-to-end model for manual-based appliance manipulation planning. Beyond the three main planning objectives, \model incorporates three complementary auxiliary tasks—bidirectional manual-part alignment, key-step action prediction, and part state judgment—that provide targeted supervision for spatial grounding, parameter precision, and state-aware plan revision, respectively. On \benchmark, \model with only 7B parameters consistently outperforms proprietary MLLMs such as GPT-5 and embodied planning baselines across all tasks: it achieves over 10$\times$ the best baseline on open-loop planning (31.36\% vs.\ 2.68\% task success) and leads the most realistic sequential setting by a wide margin (28.07\% vs.\ 4.08\%). As shown in Figure~\ref{fig:teaser}, real-robot experiments on six household appliances further confirm effective physical transfer, with 40.00\% task success compared to 3.33\% for GPT-5.

In this work, our main contributions are:
\begin{itemize}
    \item[$\bullet$] \textbf{MAGE}, a novel scalable data synthesis pipeline based on a Hierarchical Appliance Graph (HAG) representation, which automatically generates part grounding, long-horizon planning, and closed-loop recovery data at scale from appliance manuals with strategic human verification.

    \item[$\bullet$] \textbf{\dataset}, the first large-scale dataset for manual-grounded appliance manipulation planning, spanning 22 appliance categories with 89K+ part annotations, 53K+ appliance manipulation tasks, and 33K+ closed-loop adjustment steps.

    \item[$\bullet$] \textbf{\model}, the first end-to-end model for long-horizon appliance manipualtion planning, surpassing all baselines on every \benchmark track—notably over 10$\times$ on open-loop planning (31.36\% vs.\ 2.68\% task success). Real-robot experiments further confirm substantial gains (40.00\% vs.\ 3.33\% for GPT-5).
\end{itemize}

%% file: sec/2_relatedwork.tex
\section{Related Work}
\label{sec:related_work}
\vspace{-2pt}

\subsection{Appliance Manipulation}
\vspace{-2pt}
Enabling robots to operate household appliances has been explored along two directions. The first targets direct interaction with household devices~\cite{li2024manipllm,black2024pi_0,OmniManip}, but mainly handles short-horizon behaviors (\emph{e.g.}, opening a microwave door) without manual-grounded operation. The second direction explicitly uses manuals for action planning. CheckManual~\cite{checkmanual} and ApBot~\cite{apbot} show the value of manual information, but both rely on multi-stage zero-shot pipelines with cascaded external modules, incurring high latency and compounding errors. ApBot's action space is further limited to direct state transitions and incremental adjustments, precluding single-step prediction of continuous parameters such as rotation angles. Our work addresses this gap with large-scale training data containing fully parameterized atomic actions and a unified end-to-end model.

\vspace{-2pt}
\subsection{Large Models in Embodied Planning}
\vspace{-2pt}
Multimodal foundation models~\cite{flamingo,blip2,llava,GPT-4,clip} have greatly improved long-context understanding and multimodal reasoning, spurring embodied planning research. On the action execution side, SayCan~\cite{saycan} combines language models with skill affordances for grounded decision making, PaLM-E~\cite{palme} integrates multimodal perception and language modeling in a single embodied model, RT-2~\cite{rt2} casts robot actions as text tokens to enable transfer from web-scale data, and OpenVLA~\cite{openvla} makes large-scale VLA training broadly accessible. On the planning side, RoboBrain~\cite{robobrain} and RoboBrain 2.0~\cite{robobrain2} leverage embodied planning data~\cite{openxe} to improve long-horizon plan generation with stronger spatial and temporal reasoning. However, these models are evaluated primarily on generic object manipulation and fall short on manual-grounded appliance operation. The fundamental bottleneck remains the lack of large-scale, high-quality training data tailored to these capabilities.

%% file: sec/3_data_pipeline.tex
\begin{figure*}[htp]
    \centering
    \includegraphics[width=\linewidth]{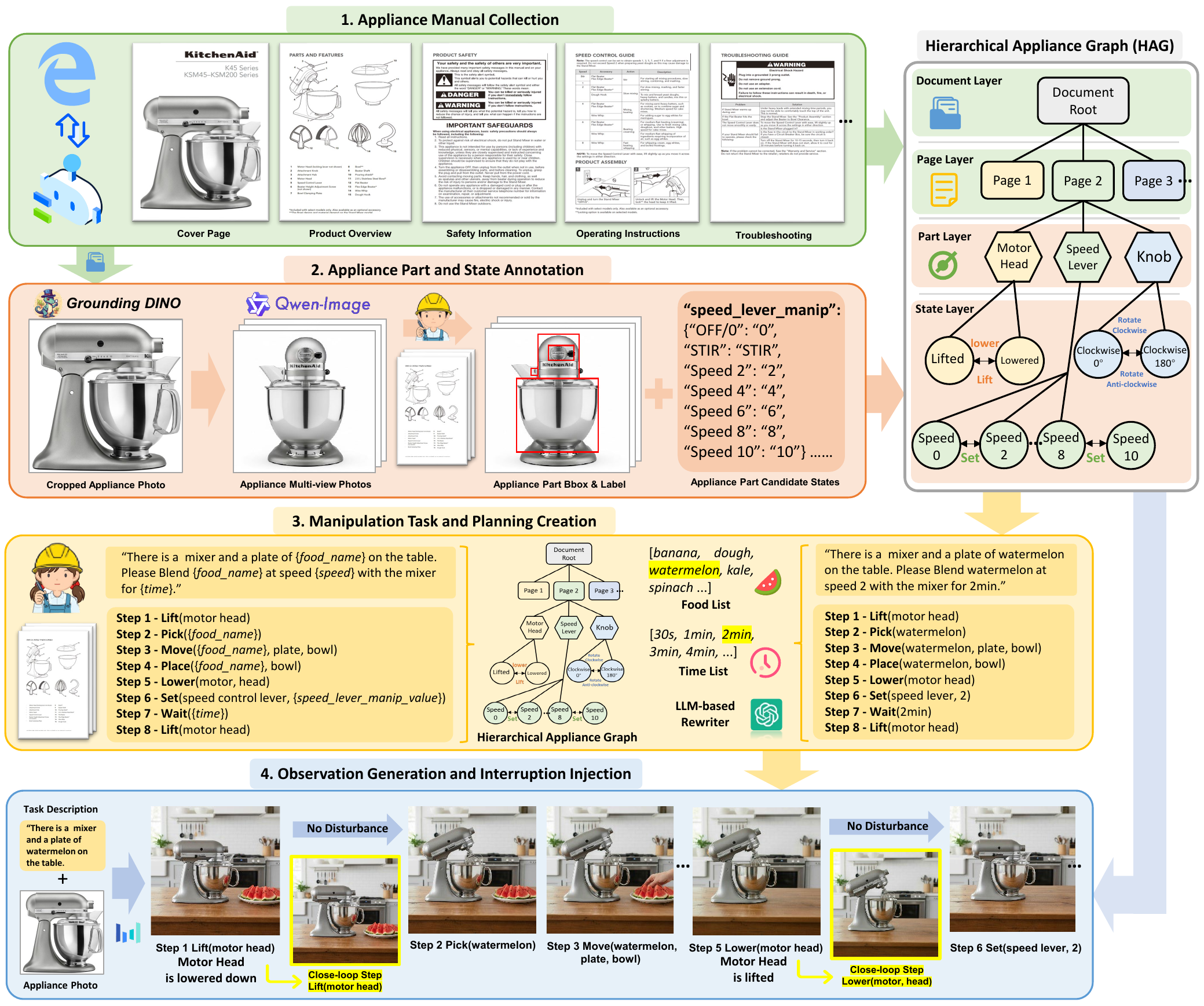}
    \caption{\textbf{Overview of MAGE.} The first two stages (Secs.~\ref{sec:stage1}--\ref{sec:stage2}) construct the Hierarchical Appliance Graph (HAG) via automated collection and human-verified annotation; the latter two (Secs.~\ref{sec:stage3}--\ref{sec:stage4}) consume the completed HAG to synthesize three types of data—part grounding annotations, open-loop manipulation plans, and  observation images with closed-loop recovery data.}
    \label{fig:applianceplan}
\end{figure*}

\section{MAGE: Data Synthesis Pipeline}
\label{sec:data_pipeline}
\vspace{-2pt}

The fundamental bottleneck for manual-grounded appliance manipulation is the lack of large-scale training data. We introduce \textbf{MAGE} (\textbf{M}anual-grounded \textbf{A}ppliance data \textbf{GE}neration pipeline), a scalable data synthesis pipeline built on a novel Hierarchical Appliance Graph (HAG). MAGE automates both HAG construction and data generation: the first two stages (Secs.~\ref{sec:stage1}--\ref{sec:stage2}) build each HAG with automated collection agents and human-verified annotation, while the latter two (Secs.~\ref{sec:stage3}--\ref{sec:stage4}) consume the completed HAG to synthesize appliance manipulation planning data with human verification gates after each stage (Figure~\ref{fig:applianceplan}).

\subsection{Hierarchical Appliance Graph Formulation}
\vspace{-2pt}
We formalize appliance manual knowledge as a \textbf{Hierarchical Appliance Graph} (HAG) $\mathcal{G}_m = (\mathcal{V}, \mathcal{E})$, where nodes $\mathcal{V}$ represent semantic entities at four granularity levels and edges $\mathcal{E}$ encode containment, reference, and state-transition relations.

A HAG organizes knowledge into four layers:
(i)~\emph{Document layer}: the manual root $v_m$ representing the entire document;
(ii)~\emph{Page layer}: typed page nodes $v_p^{(k)}$ connected to the root by containment edges, such as product-overview and operating-instructions pages;
(iii)~\emph{Part layer}: part nodes $v_c^{(j)}$ linked to their describing pages, each annotated with a visual bounding box in the corresponding appliance view;
(iv)~\emph{State layer}: state nodes $v_s^{(i)}$ representing discrete or continuous part configurations, such as knob angles and button states. Directed action edges $v_s \xrightarrow{a} v_{s'}$ encode which atomic action $a$ transitions the part from state $s$ to $s'$.

This formulation provides three useful properties:
\textbf{(1)~Combinatorial task coverage}---sampling initial--goal state pairs yields diverse tasks;
\textbf{(2)~Locally verifiable annotations}---each layer can be independently audited;
\textbf{(3)~Domain-agnostic extensibility}---the four-layer schema generalizes to any product manual.

\vspace{-2pt}
\subsection{Appliance Manual Collection}
\label{sec:stage1}
\vspace{-2pt}

\begin{figure*}[htp]
    \centering
    \includegraphics[width=\linewidth]{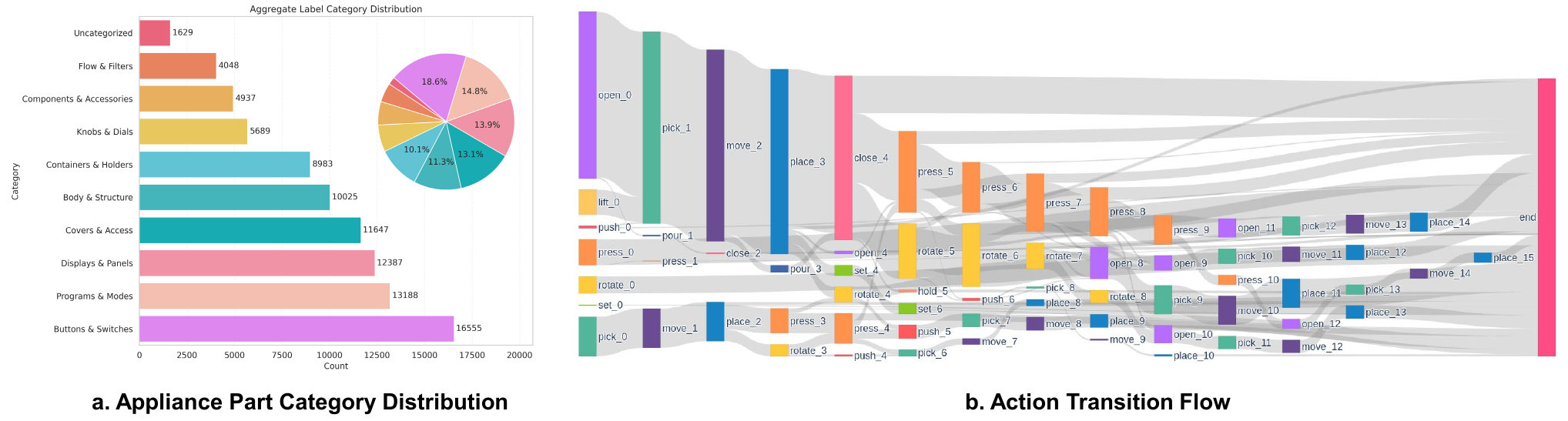}
    \caption{\textbf{Statistics of \dataset.} (a) Long-tail distribution of 89K+ part bounding boxes. (b) Action-transition flow of 53K+ open-loop planning tasks.}
    \label{fig:applianceplan_vis}
\end{figure*}

\emph{This stage builds the document and page layers of each HAG.}
We design a VLM-powered \textbf{manual collection agent} that operates in a ReAct-style loop~\cite{react} to autonomously search manufacturer websites and manual databases, download candidate manuals, and filter out low-quality ones based on page count, file size, and resolution, each instantiated as a document-layer node $v_m$. The agent then classifies each page into one of nine semantic types (product overview, operating instructions, etc.) to populate the page layer.

\vspace{-2pt}
\subsection{Appliance Part and State Annotation}
\label{sec:stage2}
\vspace{-2pt}
\emph{This stage completes the part and state layers of each HAG.}

\vspace{-2pt}
\paragraph{Part Grounding Annotation.}
Since manual diagrams and real appliance appearances differ substantially, we construct multi-view observation images to bridge this domain gap. For manuals with clear cover photos, we use an open-vocabulary object detector~\cite{liu2023grounding} with the appliance category as the prompt to detect the appliance region; if a high-confidence bounding box is returned, we crop it from the cover and then synthesize three canonical views (front, oblique front-side, and oblique top-down)~\cite{qwenimage}. Human annotators label visible parts with bounding boxes on these views guided by product-overview pages, further enriched with spatial augmentation.

\vspace{-2pt}
\paragraph{State \& Transition Annotation.}
For each part node $v_c^{(j)}$, we assign a \emph{state-type label} and construct a directed state-transition graph within the state layer. A VLM reads the relevant manual pages and proposes candidate states together with inter-state transitions; human annotators verify and correct the proposals against the manual.
All parts are unified into discrete state graphs: binary parts (\emph{e.g.}, doors) yield two-node graphs with complementary action edges, discrete parts (\emph{e.g.}, mode selectors) are enumerated by mode with button-press edges, and continuous parts (\emph{e.g.}, knobs) are discretized at manual-specified granularity with rotation edges. Edges may be unidirectional when the manual specifies irreversible operations. The resulting state nodes $\{v_s^{(i)}\}$ and directed action edges $v_s \xrightarrow{a} v_{s'}$ complete the HAG for downstream plan synthesis.

\vspace{-2pt}
\subsection{Manipualtion Task and Planning Creation}
\label{sec:stage3}
\vspace{-2pt}
\emph{This stage consumes the completed HAG to synthesize open-loop task--plan pairs, producing long-horizon manipulation planning data.}
The key idea is to treat task generation as \emph{structured graph traversal}: each task corresponds to a path in the state-transition graph, and plan generation is driven by traversing the state-transition graph and instantiating templates with sampled parameters (Figure~\ref{fig:applianceplan}).

\vspace{-2pt}
\paragraph{Task and Plan Templates.}
For each appliance, we manually design a set of task and plan templates based on its manual and the corresponding HAG. Task templates contain typed placeholders for the target state, duration, and interacting objects (\emph{e.g.}, ``\texttt{Heat the \{object\} for \{duration\} at \{target\_state\}}''). Plan templates are ordered sequences of atomic actions drawn from the action vocabulary defined in \benchmark, including appliance manipulation (\texttt{Press}, \texttt{Rotate}, etc.), object manipulation (\texttt{Pick}, \texttt{Place}, etc.), and timing (\texttt{Wait}). Every appliance action specifies the target part and an operation placeholder (rotation angle, press count, etc.); every object action specifies the target object and relevant locations; \texttt{Wait} only takes a duration parameter. Task and plan templates are paired: each task template maps to a corresponding plan template that realizes the described goal, with shared parameters kept consistent.

\vspace{-2pt}
\paragraph{Template-based Generation.}
Given a task--plan template pair, we generate concrete instances by sampling an initial--goal state pair $(s_0, s^*)$ from the HAG state layer and drawing task attributes (interacting objects, durations, etc.) from appliance-specific lists. These jointly fill the task template placeholders. The plan template already defines the action sequence skeleton; we fill operation placeholders from the state difference between $s_0$ and $s^*$ together with the sampled task attributes, yielding an open-loop plan:
\begin{equation}
a_t = \mathcal{T}_t^{\text{plan}}\!\bigl(\Delta(s_0,\, s^*),\, \mathcal{A}\bigr),\quad
Y_o = \bigl(a_1,\, a_2,\, \dots,\, a_T\bigr),
\end{equation}
where $\mathcal{T}_t^{\text{plan}}$ denotes the $t$-th step of the plan template, $\Delta(s_0, s^*)$ encodes the state difference (rotation angle $\alpha^* - \alpha_0$, discrete step count, etc.), and $\mathcal{A}$ denotes the sampled task attributes (objects, durations, etc.). An LLM rewrites the filled task templates into diverse natural-language instructions while preserving semantic fidelity. Uniform sampling of $(s_0, s^*)$ pairs ensures systematic state-space coverage, avoiding the bias of human demonstrations toward common configurations.

\vspace{-2pt}
\subsection{Observation Generation and Interruption Injection}
\label{sec:stage4}
\vspace{-2pt}
\emph{This stage generates planning-aligned observation images and injects structured disturbances to produce closed-loop recovery data.}

\vspace{-2pt}
\paragraph{Visual State Prediction.}
We use Seedream 4.0~\cite{seedream4} as the generative world model $\mathcal{M}_{\text{gen}}$ to iteratively generate observation images. Given the appliance photo and task description, we first generate the initial observation $I_0$; at each subsequent step $t$ we convert the action $a_t$ into a textual prompt describing the expected visual change and generate:
\begin{equation}
I_{t+1} = \mathcal{M}_{\text{gen}}(I_t, \texttt{prompt}(a_t)),
\end{equation}
producing an observation sequence $(I_0, I_1, \dots, I_T)$ aligned with the action sequence. We apply cross-step consistency checking and human verification to mitigate visual drift.

\vspace{-2pt}
\paragraph{Closed-loop Interruption Injection.}
To create closed-loop adjustment supervision, we inject hypothetical disturbances into intermediate steps of the generated observation sequences. We define three categories of interruptions aligned with the three part types: \emph{(i)~binary-state errors}---a part unexpectedly changes state; \emph{(ii)~button mis-touches}---an incorrect number of presses is executed; and \emph{(iii)~rotation deviations}---a knob is turned to an incorrect angle.

For each injected interruption at step $t$, we generate a post-disturbance observation image $I_t'$ using the generative world model and annotate the corrective action $a_{t+1}^{\text{corr}}$ conditioned on the task context and disturbed observation. This yields closed-loop training triples $(I_t', \text{context}, a_{t+1}^{\text{corr}})$ that teach the model to detect deviations and generate appropriate recovery actions.

%% file: sec/4_dataset.tex
\section{UseAppliance Dataset}
\label{sec:dataset}
\vspace{-2pt}

Applying the MAGE pipeline to 22 appliance categories, we build \dataset, the first large-scale dataset for manual-grounded appliance manipulation planning.

\vspace{-2pt}
\subsection{Key Characteristics}
\vspace{-2pt}
As the first large-scale training dataset for manual-grounded appliance manipulation planning, \dataset offers four key characteristics:
\begin{itemize}
    \item[$\bullet$] \emph{High quality}---All part bounding boxes and state-transition graphs are human-labeled; task and plan templates are manually authored per appliance; every annotation passes at least one round of independent human verification.
    \item[$\bullet$] \emph{Multi-task coverage}---The dataset provides unified supervision for part grounding with 89K+ bounding boxes, open-loop planning with 53K+ task--plan pairs, and closed-loop adjustment with 33K+ recovery steps, enabling joint training of all three capabilities.
    \item[$\bullet$] \emph{Rich diversity}---Spanning 22 appliance categories from manufacturers worldwide, \dataset covers a wide range of control interfaces such as knobs, buttons, touch panels, and levers, as well as diverse operation procedures including heating, blending, and brewing.
    \item[$\bullet$] \emph{Extensibility}---Since the MAGE pipeline takes only an appliance manual as input, new appliance categories can be onboarded without modifying the pipeline itself.
\end{itemize}

\vspace{-2pt}
\subsection{Data Statistics}
\vspace{-2pt}
Quantitative statistics are summarized in Table~\ref{tab:dataset_stats}. As shown in Figure~\ref{fig:applianceplan_vis}(a), part annotations follow a long-tail distribution: common parts (\emph{e.g.}, doors, buttons, knobs) receive dense annotations while appliance-specific controls appear infrequently, encouraging robustness to rare structures. Figure~\ref{fig:applianceplan_vis}(b) visualizes the action-transition flow across appliance categories: tasks average 8.15 open-loop steps, with action distributions varying significantly by category---\emph{e.g.}, knob-heavy appliances are dominated by \texttt{Rotate}, while panel-based appliances rely more on \texttt{Press}---confirming that the dataset captures diverse operation patterns.

\begin{figure*}[htp]
    \centering
    \includegraphics[width=\linewidth]{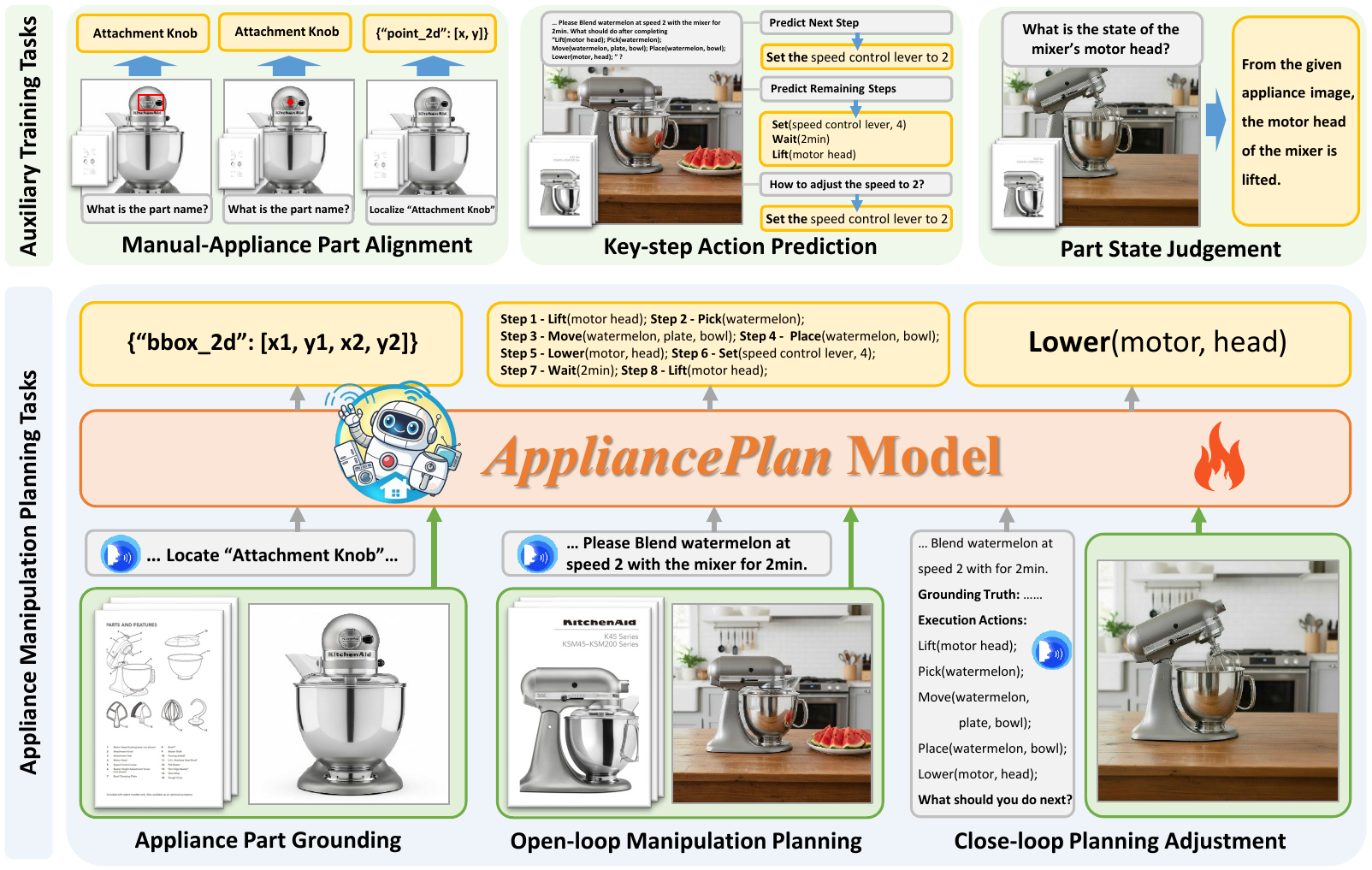}
    \caption{\textbf{Overview of \model model.} \model is an end-to-end model for manual-grounded appliance manipulation planning, trained with three main objectives and three auxiliary objectives.}
    \label{fig:model}
\end{figure*}

\input{tables/train_data_collect_Fig3}

%% file: tables/train_data_collect_Fig3.tex
\begin{table}[ht]
    \centering
    \caption{\textbf{Summary statistics of \dataset.}}
    \resizebox{\linewidth}{!}{
    \begin{tabular}{l r}
        \toprule
        \textbf{Metric} & \textbf{Value} \\
        \midrule
        Total \# Appliance manuals                        & 2,120 \\
        Total \# appliance manual pages                   & 50,727 \\
        Total \# Appliance parts                          & 1,584 \\
        Total \# Part bounding boxes                      & 89,088 \\
        Total \# Appliance manipulation tasks             & 53,854 \\
        Average \# Open-loop manipulation steps           & 8.15 \\
        Median \# Open-loop manipulation steps            & 8 \\
        Total \# Close-loop manipulation steps            & 33,815 \\
        \bottomrule
    \end{tabular}
    }
    \label{tab:dataset_stats}
\end{table}

%% file: sec/5_model.tex
\section{\model Model}
\vspace{-2pt}
\subsection{Model Architecture}
\vspace{-2pt}
\model is built on Qwen2.5-VL-7B-Instruct~\cite{qwen25vl}. Given visual inputs \(\mathcal{I}=\{I_1,\dots,I_t\}\) (manual pages and observation images) and a textual task instruction \(L\), the vision encoder and vision-language merger project the inputs into the LLM token space, and the language model predicts the output autoregressively:
\begin{equation}
P_\theta(Y\mid H_v, X)=\prod_{n=1}^{N} P_\theta\left(y_n \mid y_{<n}, H_v, X\right).
\end{equation}
where \(H_v\) denotes the projected visual tokens, \(X=\mathrm{Tok}(L)\) the text tokens, \(N\) the output sequence length, and \(Y=(y_1,\dots,y_N)\) the output sequence.

\vspace{-2pt}
\subsection{Model Training}
\vspace{-2pt}
\model is trained with 3 main objectives and 3 auxiliary objectives constructed from \dataset. All objectives are formulated as next-token prediction and optimized with cross-entropy loss.

\vspace{-2pt}
\subsubsection{Main Planning Objectives}
\vspace{-2pt}
The three main objectives correspond to the three core capabilities. Let \(\mathcal{D}_g\), \(\mathcal{D}_o\), and \(\mathcal{D}_c\) denote the training splits for part grounding, open-loop planning, and closed-loop adjustment, respectively.

\noindent\textbf{Appliance Part Grounding.}
This objective bridges the domain gap between symbolic part descriptions in manuals and their pixel-level appearance in observation images. Given a multimodal input $\mathbf{u}_g = (\mathcal{I}_{\text{manual}}, I_{\text{obs}}, L_{\text{part}})$, the model predicts the bounding-box coordinate sequence \(\mathbf{b}=(x_1,y_1,x_2,y_2)\).
\begin{equation}
\mathcal{L}_{g}=
\mathbb{E}_{(\mathbf{u}_g,\mathbf{b})\sim\mathcal{D}_g}
\left[-\log P_\theta\left(\mathbf{b}\mid \mathbf{u}_g\right)\right].
\end{equation}

\noindent\textbf{Open-loop Manipulation Planning.}
While grounding localizes \emph{where} to act, this objective trains the model to reason about \emph{what} to do and in what order. From the manual pages $\mathcal{I}_{\text{manual}}$, an initial observation $I_0$, a task instruction $L_{\text{task}}$, and the atomic action set $\mathcal{A}$, the model generates the full open-loop plan $Y_o = (a_1(\psi_1), \dots, a_T(\psi_T))$ as a single autoregressive sequence, with one parameterized action per line.
\begin{equation}
\mathcal{L}_{o}=
\mathbb{E}_{(\mathbf{u}_o,Y_o)\sim\mathcal{D}_o}
\left[-\sum_{n=1}^{|Y_o|}\log P_\theta\left(y_n^o\mid y_{<n}^o,\mathbf{u}_o\right)\right],
\end{equation}
where $\mathbf{u}_o = (\mathcal{I}_{\text{manual}}, I_0, L_{\text{task}}, \mathcal{A})$ denotes the full multimodal input.

\noindent\textbf{Closed-loop Planning Adjustment.}
This objective trains the model to predict corrective actions by comparing the expected plan with the actual execution state. At step $t$, the input comprises the current observation $I_t$, the task instruction $L_{\text{task}}$, the reference plan $P_{1:T}$, and the executed action history $A_{1:t}^{\text{exec}}$; the model predicts the next corrective action $a_{t+1}^{\star}$:
\begin{equation}
\mathcal{L}_{c}=
\mathbb{E}_{(\mathbf{u}_c,a_{t+1}^\star)\sim\mathcal{D}_c}
\left[-\log P_\theta\left(a_{t+1}^\star\mid \mathbf{u}_c\right)\right],
\end{equation}
where $\mathbf{u}_c = (I_t, L_{\text{task}}, P_{1:T}, A_{1:t}^{\text{exec}})$.

The overall main objective combines these three losses:
\begin{equation}
\mathcal{L}_{\text{main}}=
\mathcal{L}_g+\mathcal{L}_o+\mathcal{L}_c.
\end{equation}

\vspace{-2pt}
\subsubsection{Auxiliary Training Objectives}
\vspace{-2pt}
The three auxiliary objectives target complementary dimensions of the planning problem. Let \(\mathcal{D}_{\text{align}}\), \(\mathcal{D}_{\text{key}}\), and \(\mathcal{D}_{\text{state}}\) denote the corresponding training splits.

\noindent\textbf{Manual-Appliance Part Alignment.}
The main grounding objective only maps part names to locations. This auxiliary objective strengthens bidirectional part--manual correspondence with three complementary sub-objectives, all receiving manual pages and an observation image as input:
\emph{(i)~Manual-side part identification}---the target part is highlighted by a bounding box $\mathbf{b}$ on the observation image; the model predicts the part name $n^\star$;
\emph{(ii)~Observation-side part identification}---the target part is indicated by its center-point coordinate $\mathbf{c}$; the model predicts $n^\star$;
\emph{(iii)~Part localization}---a part name $L_{\text{part}}$ is provided; the model predicts the center-point coordinate $\mathbf{c}$ on the observation image.
The alignment loss is:
\begin{equation}
\mathcal{L}_{\text{align}}\!=\!
\mathbb{E}_{\mathcal{D}_{\text{align}}}\!\left[-\log P_\theta(n^\star\!\mid\!\mathbf{u}_{\text{mid}})
\!-\!\log P_\theta(n^\star\!\mid\!\mathbf{u}_{\text{oid}})
\!-\!\log P_\theta(\mathbf{c}\!\mid\!\mathbf{u}_{\text{loc}})\right]\!,
\end{equation}
where $\mathbf{u}_{\text{mid}}\!=\!(\mathcal{I}_{\text{manual}}, I_{\text{obs}}, \mathbf{b})$,\, $\mathbf{u}_{\text{oid}}\!=\!(\mathcal{I}_{\text{manual}}, I_{\text{obs}}, \mathbf{c})$, and $\mathbf{u}_{\text{loc}}\!=\!(\mathcal{I}_{\text{manual}}, I_{\text{obs}}, L_{\text{part}})$.

\noindent\textbf{Key-Step Action Prediction.}
This objective focuses supervision on steps involving precise action parameter prediction (\emph{e.g.}, rotation angles, button-press counts). We extract such steps from open-loop plans to form a key-step subset $\mathcal{D}_{\text{key}}$ and train the model with three complementary prediction targets:
\emph{(i)~Next-step prediction}---from manual pages, the current observation, the task instruction, and the executed action history, the model predicts the next key action;
\emph{(ii)~Remaining-step prediction}---with the same context, predict all remaining actions from the current step;
\emph{(iii)~Parameter VQA}---the model answers a parameter question (\emph{e.g.}, target angle, press count) based on the current observation.
\begin{equation}
\mathcal{L}_{\text{key}}=
\mathbb{E}_{(\mathbf{u}_{\text{key}},a_{\text{key}}^\star)\sim\mathcal{D}_{\text{key}}}
\left[-\log P_\theta\left(a_{\text{key}}^\star\mid \mathbf{u}_{\text{key}}\right)\right],
\end{equation}
where $\mathbf{u}_{\text{key}} = (\mathcal{I}_{\text{manual}}, I_t, L_{\text{task}}, A_{1:t})$.

\noindent\textbf{Part State Judgment.}
This objective provides explicit state-recognition supervision through VQA: the model receives manual pages, an observation image $I_{\text{obs}}$, and a question about a specific part's state, and predicts the ground-truth state label $s^\star$.
\begin{equation}
\mathcal{L}_{\text{state}}=
\mathbb{E}_{(\mathbf{u}_{\text{state}},s^\star)\sim\mathcal{D}_{\text{state}}}
\left[-\log P_\theta\left(s^\star\mid \mathbf{u}_{\text{state}}\right)\right],
\end{equation}
where $\mathbf{u}_{\text{state}} = (\mathcal{I}_{\text{manual}}, I_{\text{obs}}, L_{\text{state\_query}})$ and $L_{\text{state\_query}}$ is the state question.

The overall auxiliary objective is:
\begin{equation}
\mathcal{L}_{\text{aux}}=
\mathcal{L}_{\text{align}}+
\mathcal{L}_{\text{key}}+
\mathcal{L}_{\text{state}}.
\end{equation}
The full training objective is:
\begin{equation}
\mathcal{L}=\mathcal{L}_{\text{main}}+\mathcal{L}_{\text{aux}}.
\end{equation}

%% file: sec/6_experiments.tex
\section{Experiments}
\vspace{-2pt}

\input{tables/single_task}

\vspace{-2pt}
\subsection{Implementation Details}
\vspace{-2pt}
Training runs for 96 GPU-hours on NVIDIA H200 with a global batch size of 512, a learning rate of \(1\times10^{-5}\) with cosine decay, and a frozen vision encoder. We use DeepSpeed ZeRO-3~\cite{deepspeed} with bfloat16 mixed precision. All loss weights are set to 1.0. During inference, we use a temperature of 0.1 for stable plan generation. Training uses only \dataset, which has no overlap with \benchmark, making all evaluations out-of-distribution.

\input{tables/sequential_plan}

\vspace{-2pt}
\subsection{Evaluation on \benchmark}
\vspace{-2pt}
\subsubsection{Evaluation Tracks and Metrics}
\vspace{-2pt}

\benchmark~\cite{realappliance} is a benchmark for manual-grounded appliance manipulation built on 100 high-fidelity digital appliance assets in NVIDIA Isaac Sim, spanning 14 common appliance categories. Each asset is aligned with its real-world manual in appearance, physical mechanisms, electronic mechanisms, and program logic. We consider the following four evaluation tracks:
\begin{itemize}
    \item[$\bullet$] \textbf{\textit{Appliance Part Grounding}}: Given relevant manual pages, observation, and part name, the model localizes the target appliance part by predicting its bounding box. We report average Intersection-over-Union (IoU) and mean Average Precision at IoU 0.5 (mAP@0.5).

    \item[$\bullet$] \textbf{\textit{Open-loop Manipulation Planning}}: Given relevant manual pages, an initial observation image, a task instruction, and the available atomic action set, the model predicts the full manipulation sequence without feedback during execution. Following the benchmark protocol, we report \emph{task completion rate} and \emph{task success rate}.

    \item[$\bullet$] \textbf{\textit{Closed-loop Planning Adjustment}}: Given relevant manual pages, a task instruction, a reference plan, the executed action history, the current observation image, and the available atomic action set, the model predicts the next corrective action under disturbances. Performance is measured by \emph{step-wise success rate}.

    \item[$\bullet$] \textbf{\textit{Sequential Planning and Adjustment}}: This evaluation-only setting combines the previous two tracks: the model first generates an open-loop plan from the manual, initial observation, task instruction, and available atomic action set, and then revises subsequent actions online from execution history and updated observations. We use \emph{task completion rate} and \emph{task success rate} as metrics.
\end{itemize}
The first three tracks are defined by \benchmark. The fourth track based on \benchmark is introduced by us.

\begin{figure}[H]
    \centering
    \includegraphics[width=\linewidth]{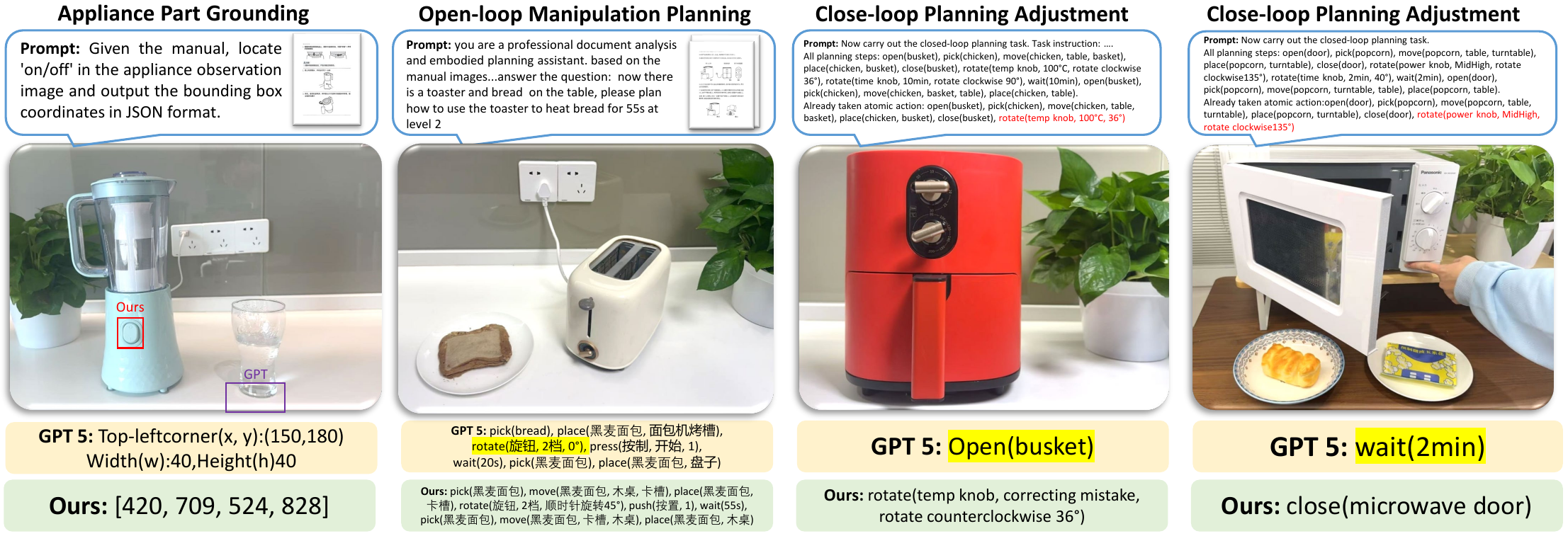}
    \caption{\textbf{Real-robot execution with \model.} Representative examples on household appliances.}
    \label{fig:realcase}
\end{figure}

\vspace{-2pt}
\subsubsection{Comparison with Baselines}
\vspace{-2pt}
Table~\ref{tab:single_task} compares \model with proprietary MLLMs, open-source MLLMs, and embodied planning baselines.

\vspace{-2pt}
\paragraph{Appliance Part Grounding}
\model achieves the best overall grounding performance (22.96\% IoU, 22.24\% mAP@0.5), outperforming the strongest embodied baseline ApBot by 12.36 IoU and 10.14 mAP@0.5 points with a simpler end-to-end pipeline. The gains stem from explicit manual-conditioned training: without manual context, models cannot reliably associate part descriptions with visual appearance, especially for appliance-specific controls absent from web-crawled data.

\vspace{-2pt}
\paragraph{Open-loop Manipulation Planning}
\model achieves 47.86\% task completion and 31.36\% task success—over 10$\times$ the best baseline (4.36\%/2.68\%)—and ranks first on every appliance category. The near-zero baseline performance confirms that general language and vision capabilities do not transfer to structured multi-step planning with precise parameters; training on (initial-state, goal-state, plan) triples with fully parameterized atomic actions bridges this gap.

\vspace{-2pt}
\paragraph{Closed-loop Planning Adjustment}
\model achieves the highest step-wise success rate (37.12\%), outperforming the best open-source baseline RoboBrain 2.0-7B by 5.35 points and the best proprietary MLLM Gemini 2.5 Pro by 5.39 points, indicating stronger robustness to execution deviations across appliance categories.

\vspace{-2pt}
\paragraph{Sequential Planning and Adjustment}
This end-to-end setting is the most realistic, as both the initial plan and subsequent corrections are model-generated. \model achieves 44.59\% completion and 28.07\% success, versus 5.84\%/4.08\% for the best baseline (Qwen3-VL 235B Instruct), confirming that planning and correction gains transfer to long-horizon execution, including recovery from errors introduced by the model's own predictions.

\vspace{-2pt}
\subsubsection{Ablation of Auxiliary Training Objectives}
\vspace{-2pt}
We ablate the three auxiliary objectives by removing one at a time; results appear in the lower rows of Table~\ref{tab:single_task} (rows marked ``$-$'').
Removing \textbf{Manual-Appliance Part Alignment} causes the largest grounding degradation, confirming that bidirectional part--manual correspondence is essential for accurate spatial localization. Removing \textbf{Key-Step Action Prediction} produces the largest decline on long-horizon metrics—particularly on the realistic sequential setting—indicating that focused supervision on parameter-critical steps is the primary driver of precise action prediction. Removing \textbf{Part State Judgment} mainly harms closed-loop recovery, validating its role in enabling state-aware plan revision.
Overall, each objective contributes to its targeted capability as designed, and their combination yields the best balance across grounding, planning, and closed-loop correction.

\vspace{-2pt}
\subsection{Real-Robot Evaluation}
\vspace{-2pt}
To test whether benchmark gains transfer to physical execution, we deploy \model on a local workstation with an RTX 4090 GPU and evaluate it on a Franka Emika Panda robot. The system uses an eye-in-hand RealSense D415 camera together with an third-view D415 camera. We test six household appliances: microwave, air fryer, toaster, rice cooker, coffee machine, and blender. For each appliance, we build a CAD/URDF model; FoundationPose~\cite{foundationpose} uses the CAD model for 6D pose estimation, from which appliance-specific atomic actions are instantiated. For \emph{Pick} actions, we integrate AnyGrasp~\cite{anygrasp}.
During each episode, \model first predicts an open-loop plan from the manual, observation, and task instruction before execution, and then revises the next action online using real-time perception feedback. We evaluate 10 manipulation tasks for each appliance. Quantitative results are reported in Table~\ref{tab:real_robot}, and qualitative examples are shown in Figure~\ref{fig:realcase}. As shown in Table~\ref{tab:real_robot}, \model outperforms all baselines on every appliance. Averaged over the six appliances, \model achieves 49.87\% task completion and 40.00\% task success, compared with 19.35\%/3.33\% for GPT-5 and 9.12\%/3.33\% for ApBot. The much larger margin on task success indicates that general-purpose MLLMs and pipeline methods may execute early steps correctly, but often fail at the state-dependent decisions required to complete the full task.

\input{tables/real_robot}

%% file: tables/single_task.tex
\begin{table*}[tb]
    \begin{center}
    \small
    \caption{\textbf{Results on \benchmark across three standard evaluation tracks.}} 
    \resizebox{\textwidth}{!}{
        \setlength{\tabcolsep}{1.0mm}{
        \begin{tabular}{r|c|c|c|c|c|c|c|c|c|c|c|c|c|c|c}
  \toprule
    \multirow{1}{*}{\centering \textbf{\textsc{Baseline Model}}}
    & \includegraphics[width=0.025\linewidth]{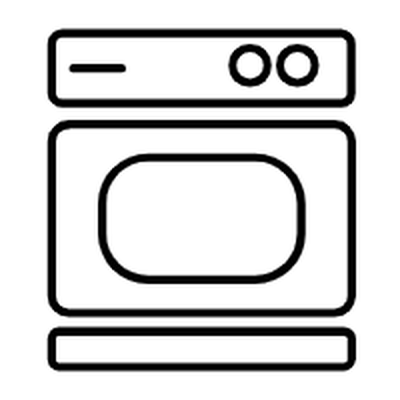}
    & \includegraphics[width=0.025\linewidth]{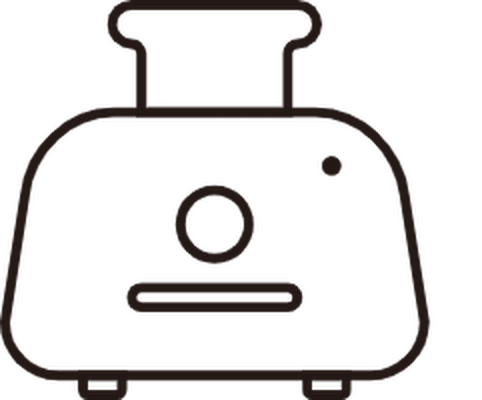}
    & \includegraphics[width=0.025\linewidth]{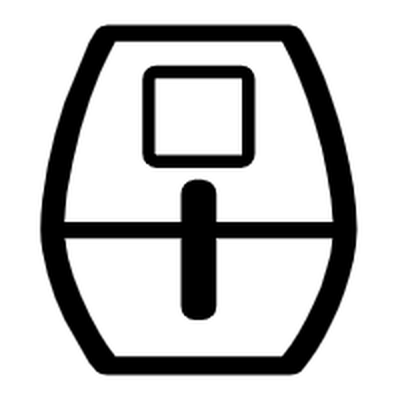}
    & \includegraphics[width=0.025\linewidth]{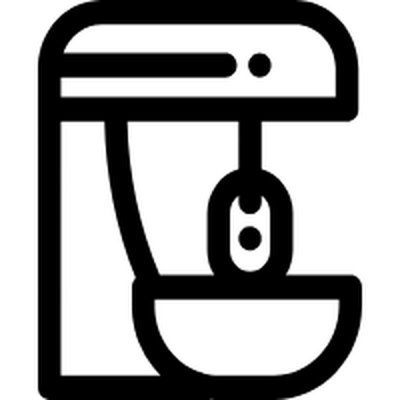}
    & \includegraphics[width=0.025\linewidth]{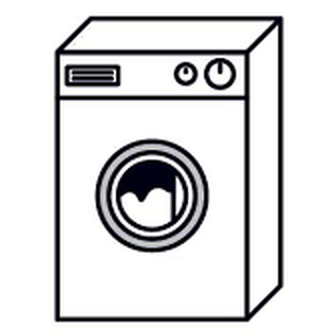}
    & \includegraphics[width=0.025\linewidth]{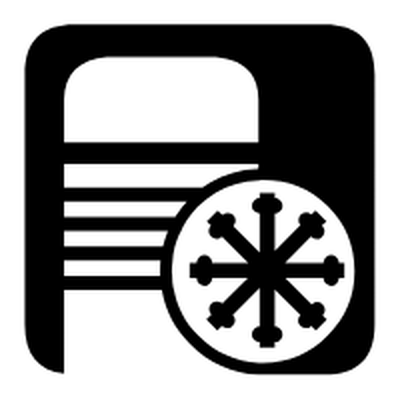}
    & \includegraphics[width=0.025\linewidth]{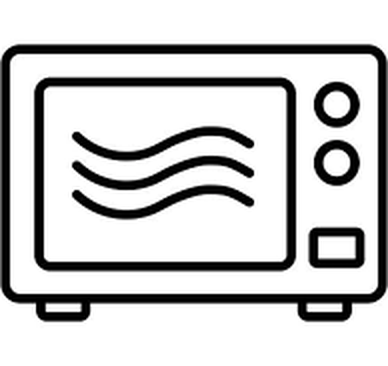}
    & \includegraphics[width=0.025\linewidth]{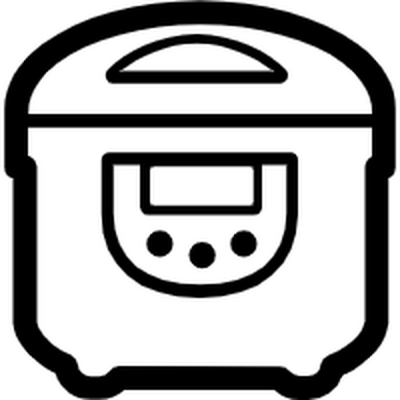}
    & \includegraphics[width=0.025\linewidth]{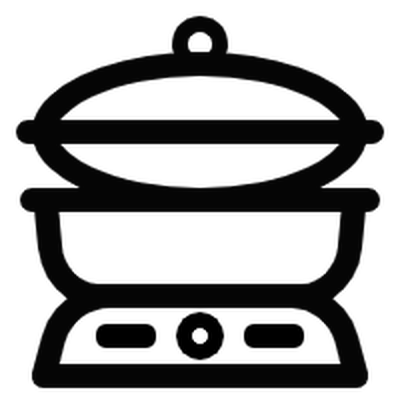}
    & \includegraphics[width=0.025\linewidth]{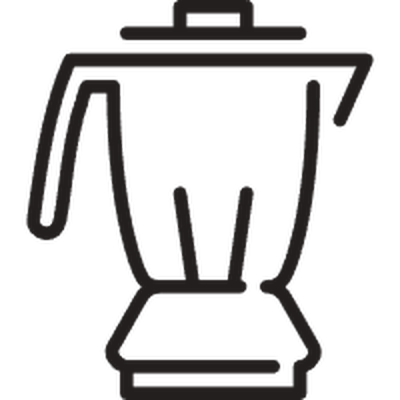}
    & \includegraphics[width=0.025\linewidth]{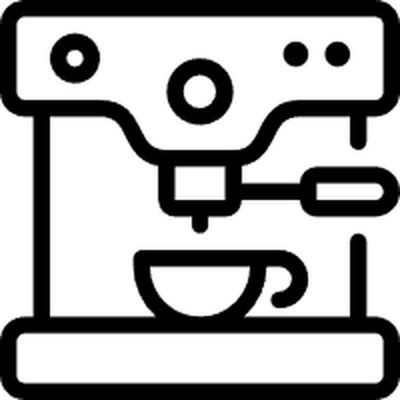}
    & \includegraphics[width=0.025\linewidth]{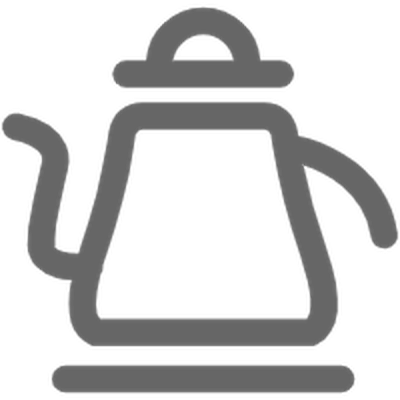}
    & \includegraphics[width=0.025\linewidth]{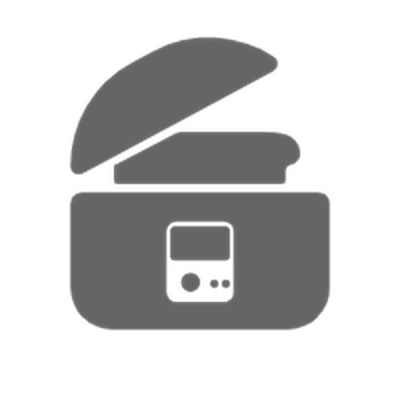}
    & \includegraphics[width=0.025\linewidth]{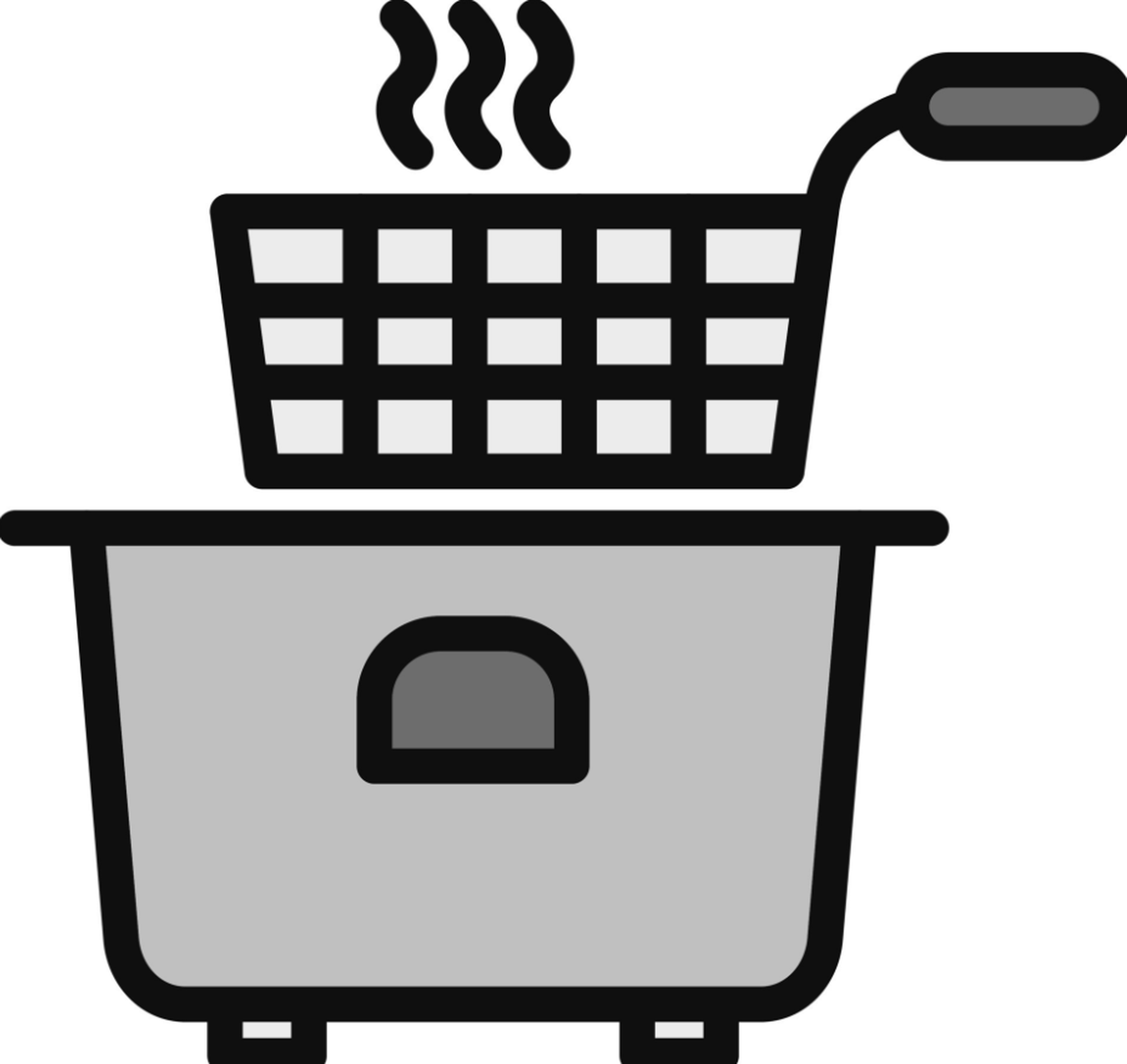}
    & \textbf{Total} \\ 
  \toprule
  \multicolumn{15}{c}{\textbf{Appliance Part Grounding} (\emph{Average / mAP@0.5})} \\
  \midrule
  \rowcolor{blue!8} \multicolumn{16}{l}{\textit{Proprietary MLLMs}} \\
  GPT-5~\cite{GPT-5}  &12.71/7.69 &4.84/0.00 &12.00/8.51 &16.33/17.85 &13.23/9.67 &\textbf{28.83/28.57} &\textbf{20.24}/11.94 &10.42/4.34 &32.97/33.33 &10.01/9.67 &4.72/1.33 &\textbf{19.17/18.75} &11.17/9.67 &4.80/0.00 &12.15/8.59 \\ 
  GPT-5 Mini~\cite{GPT-5}  &9.08/7.46 &1.85/0.00 &7.18/3.19 &9.71/0.00 &4.43/3.22 &10.83/0.00 &13.95/14.92 &2.45/0.00 &19.56/0.00 &1.82/0.00 &2.48/0.00 &7.59/6.25 &5.33/0.00 &4.94/0.00 &6.51/3.49 \\ 
  Gemini 2.5 Pro~\cite{gemini}  &7.77/4.47 &3.75/0.00 &10.74/8.51 &9.47/10.71 &6.74/6.45 &7.23/0.00 &10.46/8.95 &5.87/4.34 &32.82/41.66 &9.70/11.29 &2.92/0.00 &6.64/6.25 &6.70/6.45 &6.16/0.00 &8.16/6.64 \\ 
  Gemini 2.5 Flash~\cite{gemini}  &8.21/4.47 &2.46/0.00 &7.94/7.44 &12.14/17.85 &2.28/3.22 &16.67/0.00 &11.22/7.46 &2.18/0.00 &21.16/16.66 &6.06/4.83 &1.52/1.33 &5.06/0.00 &4.81/6.45 &8.38/0.00 &6.67/5.06 \\ 
  \rowcolor{blue!8} \multicolumn{16}{l}{\textit{Open-source MLLMs}} \\
  Qwen3-VL 8B Thinking~\cite{Qwen3-vl}  &5.55/1.49 &0.25/0.00 &2.14/1.06 &2.02/0.00 &1.11/0.00 &5.29/0.00 &6.09/0.00 &3.93/0.00 &5.25/0.00 &2.52/1.61 &0.40/0.00 &0.00/0.00 &3.03/3.22 &9.93/0.00 &2.92/0.69 \\
  Qwen3-VL 8B Instruct~\cite{Qwen3-vl}  &2.27/0.00 &0.00/0.00 &0.44/0.00 &0.63/0.00 &1.28/0.00 &0.15/0.00 &2.83/0.00 &1.56/0.00 &4.58/0.00 &0.85/0.00 &0.32/0.00 &0.00/0.00 &1.46/0.00 &8.32/0.00 &1.32/0.00 \\
  Qwen3-VL 235B Thinking~\cite{Qwen3-vl}  &5.19/2.98 &0.99/0.00 &3.41/3.19 &1.01/0.00 &1.96/0.00 &0.25/0.00 &6.78/0.00 &1.58/0.00 &6.01/0.00 &1.08/0.00 &0.12/0.00 &0.00/0.00 &1.50/0.00 &8.64/0.00 &2.80/0.87 \\
  Qwen3-VL 235B Instruct~\cite{Qwen3-vl} &4.58/4.47 &0.00/0.00 &0.61/0.00 &0.61/0.00 &1.15/0.00 &0.46/0.00 &4.64/0.00 &1.61/0.00 &6.28/0.00 &0.86/0.00 &0.00/0.00 &0.00/0.00 &0.44/0.00 &7.35/0.00 &1.75/0.52 \\
  GLM-4.1V-Thinking~\cite{glm}  &3.88/1.49 &0.00/0.00 &0.54/0.00 &0.98/0.00 &1.67/0.00 &0.42/0.00 &4.01/0.00 &1.55/0.00 &4.71/0.00 &0.93/0.00 &0.00/0.00 &0.00/0.00 &1.33/0.00 &11.63/0.00 &1.74/0.17 \\
  GLM-4.5V~\cite{glm}  &3.59/2.98 &0.00/0.00 &0.75/0.00 &0.99/0.00 &2.50/0.00 &0.32/0.00 &4.09/0.00 &1.56/0.00 &6.15/0.00 &0.66/0.00 &0.00/0.00 &0.00/0.00 &0.62/0.00 &11.00/0.00 &1.76/0.35 \\
  \rowcolor{blue!8} \multicolumn{16}{l}{\textit{Embodied Planning Models}} \\
  RoboBrain 2.0-7B~\cite{robobrain2}  &0.00/0.00 &0.00/0.00 &0.00/0.00 &0.00/0.00 &0.00/0.00 &0.00/0.00 &0.00/0.00 &0.00/0.00 &0.00/0.00 &0.00/0.00 &0.00/0.00 &0.00/0.00 &0.00/0.00 &0.00/0.00 &0.00/0.00 \\
  RoboBrain 2.0-32B~\cite{robobrain2}  &0.00/0.00 &0.00/0.00 &0.00/0.00 &0.00/0.00 &0.00/0.00 &0.00/0.00 &0.00/0.00 &0.00/0.00 &0.00/0.00 &0.00/0.00 &0.00/0.00 &0.00/0.00 &0.00/0.00 &0.00/0.00 &0.00/0.00 \\
  ManualPlan~\cite{checkmanual}  &3.56/0.00 &0.00/0.00 &1.01/0.00 &2.46/0.00 &2.37/0.00 &3.08/0.00 &5.61/0.00 &1.25/0.00 &2.79/0.00 &0.20/0.00 &0.15/0.00 &1.66/0.00 &1.73/0.00 &5.07/0.00 &1.92/0.00 \\
  ApBot~\cite{apbot}  &\textbf{23.60/28.40} &10.50/12.30 &8.50/7.60 &4.70/4.50 &\textbf{15.70/20.00} &17.70/20.00 &10.20/11.20 &0.00/0.00 &4.50/4.30 &11.70/12.00 &11.20/14.70 &0.00/0.00 &12.20/13.60 &4.30/5.30 &10.60/12.10 \\
  \rowcolor{gray!10}
  \model (Ours)  &13.25/10.45& \textbf{19.03/19.15}& \textbf{14.19/12.77}& \textbf{35.04/35.71}& 15.25/16.13& 16.39/14.29& 19.77/\textbf{19.70}& \textbf{39.65/39.13}& \textbf{72.06/66.67}& \textbf{27.82/27.42}& \textbf{31.98/33.33}& 6.77/6.25& \textbf{20.53/19.35}& \textbf{39.56/33.33}& \textbf{22.96/22.24} \\
  \rowcolor{gray!10}
  - \textit{Manual-App. Part Alignment} & 17.74/11.94 & 5.77/2.13 & 13.42/10.64 & 18.09/21.43 & 22.43/19.35 & 25.78/14.29 & 23.75/21.21 & 19.56/21.74 & 52.06/58.33 & 18.46/14.52 & 26.93/32.00 & 16.63/18.75 & 14.04/6.45 & 32.73/25.00 & 19.27/17.34 \\
  \rowcolor{gray!10}
  - \textit{Key-step Action Prediction} &26.66/26.87 & 14.50/12.77 & 15.26/11.70 & 27.57/28.57 & 32.56/29.03 & 29.50/42.86 & 25.56/21.21 & 22.94/26.09 & 50.59/58.33 & 21.41/22.58 & 26.35/28.00 & 22.39/18.75 & 14.45/9.68 & 25.44/16.67 & 22.99/21.89 \\
  \rowcolor{gray!10}
  - \textit{Part State Judgement} & 24.62/22.39 & 19.69/19.15 & 18.34/15.96 & 24.41/25.00 & 33.44/32.26 & 29.31/42.86 & 29.00/27.27 & 22.98/26.09 & 41.72/33.33 & 18.64/16.13 & 26.59/32.00 & 33.08/31.25 & 9.50/3.23 & 20.89/16.67 & 23.45/22.59 \\
  \midrule
  \multicolumn{15}{c}{\textbf{Open-loop Manipulation Planning} (\emph{Task Completion Rate / Task Success Rate})} \\ 
  \midrule
  \rowcolor{blue!8} \multicolumn{16}{l}{\textit{Proprietary MLLMs}} \\ 
  GPT-5~\cite{GPT-5}  &2.60/1.11 &2.57/0.00 &1.55/0.00 &0.00/0.00 &7.77/2.00 &5.92/0.00 &4.70/0.83 &0.95/0.00 &4.80/2.00 &3.02/0.00 &6.44/2.02 &7.77/0.00 &14.59/6.00 &10.66/10.00 &4.30/1.22  \\ 
  GPT-5 Mini~\cite{GPT-5}  &2.22/2.22 &5.23/1.25 &0.11/0.00 &0.00/0.00 &3.58/0.00 &1.26/0.00 &3.52/1.66 &6.45/0.00 &4.09/2.00 &2.34/0.00 &4.08/0.00 &5.55/0.00 &9.29/4.00 &9.66/6.66 &3.27/1.02 \\ 
  Gemini 2.5 Pro~\cite{gemini}  &0.59/0.00 &3.47/1.25 &3.70/1.53 &0.00/0.00 &3.18/0.00 &2.01/0.00 &4.51/4.16 &0.47/0.00 &0.50/0.00 &1.07/0.00 &10.72/8.08 &8.88/0.00 &15.80/12.00 &8.00/6.66 &4.08/2.45 \\ 
  Gemini 2.5 Flash~\cite{gemini}  &0.23/0.00 &2.00/0.00 &6.12/3.84 &0.00/0.00 &2.79/0.00 &1.26/0.00 &2.34/1.66 &7.82/3.33 &2.16/2.00 &2.34/0.88 &10.55/8.08 &5.55/0.00 &15.51/12.00 &6.66/6.66 &4.26/2.65 \\ 
  \rowcolor{blue!8} \multicolumn{16}{l}{\textit{Open-source MLLMs}} \\
  Qwen3-VL 8B Thinking~\cite{Qwen3-vl}  &1.55/1.11 &2.82/0.00 &1.53/1.53 &0.00/0.00 &3.88/2.00 &4.23/3.70 &1.79/1.66 &7.42/3.33 &5.73/2.00 &0.88/0.88 &6.39/5.05 &2.77/0.00 &7.22/4.00 &6.66/6.66 &3.01/1.94 \\
  Qwen3-VL 8B Instruct~\cite{Qwen3-vl}  &0.00/0.00 &4.41/0.00 &0.76/0.76 &0.00/0.00 &0.00/0.00 &0.00/0.00 &1.66/1.66 &0.00/0.00 &2.00/2.00 &0.88/0.88 &1.68/0.00 &12.22/0.00 &6.71/4.65 &3.33/3.33 &1.70/0.82 \\
  Qwen3-VL 235B Thinking~\cite{Qwen3-vl}   &1.56/1.11 &5.95/1.25 &1.59/0.00 &0.00/0.00 &9.46/4.00 &7.83/3.70 &2.08/1.66 &10.79/3.33 &5.40/2.00 &4.18/0.88 &6.68/4.04 &8.33/0.00 &7.51/2.00 &8.00/6.66 &4.36/1.73 \\
  Qwen3-VL 235B Instruct~\cite{Qwen3-vl}   &0.00/0.00 &6.86/1.25 &3.23/3.07 &0.00/0.00 &8.01/4.00 &2.32/0.00 &4.50/4.16 &3.14/0.00 &3.13/2.00 &2.47/1.76 &6.58/3.03 &8.88/0.00 &9.95/6.00 &6.66/6.66 &4.11/2.34 \\
  GLM-4.1V-Thinking~\cite{glm}  &0.00/0.00 &0.00/0.00 &0.00/0.00 &0.00/0.00 &0.00/0.00 &0.00/0.00 &0.00/0.00 &0.00/0.00 &0.00/0.00 &0.00/0.00 &0.00/0.00 &0.00/0.00 &0.00/0.00 &0.00/0.00 &0.00/0.00 \\
  GLM-4.5V~\cite{glm}  &2.22/2.22 &3.30/1.25 &2.41/2.30 &1.11/1.11 &5.21/4.00 &4.44/3.70 &1.30/0.83 &1.33/0.00 &4.43/2.00 &1.50/0.97 &8.41/6.06 &3.33/0.00 &10.11/8.00 &12.66/10.00 &3.73/2.68 \\
  \rowcolor{blue!8} \multicolumn{16}{l}{\textit{Embodied Planning Models}} \\
  RoboBrain 2.0-7B~\cite{robobrain2}  &0.00/0.00 &0.84/0.00 &0.06/0.00 &0.00/0.00 &0.17/0.00 &0.00/0.00 &0.16/0.00 &0.00/0.00 &0.00/0.00 &0.27/0.00 &0.00/0.00 &0.55/0.00 &0.22/0.00 &0.00/0.00 &0.16/0.00 \\
  RoboBrain 2.0-32B~\cite{robobrain2}  &0.00/0.00 &1.99/0.00 &0.00/0.00 &0.00/0.00 &0.71/0.00 &0.00/0.00 &0.12/0.00 &0.00/0.00 &0.00/0.00 &0.17/0.00 &0.67/0.00 &0.00/0.00 &1.53/0.00 &0.00/0.00 &0.37/0.00 \\
  ManualPlan~\cite{checkmanual}  &6.12/0.00 &8.76/0.00 &2.003/0.00 &0.92/0.00 &3.20/0.00 &5.75/0.00 &11.27/0.00 &10.44/0.00 &6.44/0.00 &4.51/1.76 &2.20/0.00 &11.32/0.00 &11.50/3.99 &2.17/0.00 &5.61/0.40 \\
  ApBot~\cite{apbot}  &0.00/0.00 &0.00/0.00 &0.00/0.00 &0.00/0.00 &2.00/2.00 &8.70/7.40 &1.60/0.80 &0.00/0.00 &0.00/0.00 &7.10/7.10 &7.10/7.10 &0.00/0.00 &4.40/4.00 &0.00/0.00 &2.30/2.10\\
  \rowcolor{gray!10}
  \model (Ours) & \textbf{45.70/27.80} & \textbf{42.80/23.70} & \textbf{48.00/36.20} & \textbf{49.00/21.10} & \textbf{29.92/14.00} & \textbf{43.40/40.70} & \textbf{48.20/31.70} & \textbf{20.00/13.30} & \textbf{41.58/24.02} & \textbf{47.30/32.70} & \textbf{61.30/49.50} & \textbf{73.30/60.00} & \textbf{61.82/45.98} & \textbf{52.20/13.30} & \textbf{47.86/31.36} \\
\rowcolor{gray!10}

-\textit{Manual-App. Part Alignment} & 45.50/28.90 & 34.30/16.30 & 45.00/34.60 & 48.70/24.40 & 36.68/15.98 & 63.40/55.60 & 36.60/15.80 & 28.70/6.70 & 43.22/23.98 & 53.10/41.60 & 52.80/41.40 & 100.00/100.00 & 66.26/43.98 & 53.70/16.70 & 47.17/30.33 \\
\rowcolor{gray!10}

-\textit{Key-step Action Prediction} &36.10/17.80 & 34.40/17.50 & 32.10/21.50 & 46.70/18.90 & 20.68/12.02 & 42.20/40.70 & 36.50/17.50 & 23.40/6.70 & 38.94/14.02 & 53.60/34.50 & 46.60/20.20 & 83.30/70.00 & 69.26/42.02 & 54.60/16.70 & 41.90/22.57
 \\
\rowcolor{gray!10}

-\textit{Part State Judgement} & 45.90/26.70 & 34.00/15.00 & 33.00/25.40 & 54.30/33.30 & 34.70/18.02 & 53.30/51.90 & 41.30/20.80 & 8.70/6.70 & 26.28/22.02 & 53.50/41.60 & 51.00/34.30 & 95.00/95.00 & 49.86/27.98 & 51.20/16.70 & 43.67/28.50 \\
  \midrule
  \multicolumn{15}{c}{\textbf{Close-loop Planning Adjustment} (\emph{Step-wise Success Rate})} \\
  \midrule
  \rowcolor{blue!8} \multicolumn{16}{l}{\textit{Proprietary MLLMs}} \\ 
  GPT-5~\cite{GPT-5}  &28.84 &5.79 &38.51 &0.00 &31.74 &12.50 &39.43 &6.66 &\textbf{41.66} &30.55 &20.58 &10.52 &34.14 &\textbf{45.45} &29.61 \\ 
  GPT-5 Mini~\cite{GPT-5}  &14.42 &5.17 &18.24 &0.00 &15.78 &3.84 &25.19 &15.00 &37.83 &9.09 &20.58 &\textbf{42.10} &0.00 &27.27 &16.33 \\ 
  Gemini 2.5 Pro~\cite{gemini}  &26.13 &13.04 &\textbf{45.62} &0.00 &42.85 &11.53 &38.00 &5.00 &27.02 &35.45 &23.52 &15.78 &\textbf{41.46} &31.81 &31.73 \\ 
  Gemini 2.5 Flash~\cite{gemini}  &31.06 &10.14 &34.05 &0.00 &46.03 &19.23 &37.90 &15.00 &24.32 &37.03 &26.47 &\textbf{42.10} &39.02 &33.33 &31.61 \\ 
  \rowcolor{blue!8} \multicolumn{16}{l}{\textit{Open-source MLLMs}} \\
  Qwen3-VL 8B Thinking~\cite{Qwen3-vl}  &23.07 &13.04 &30.43 &0.00 &30.15 &11.53 &36.36 &15.00 &13.51 &37.27 &20.58 &36.84 &0.00 &22.72 &25.58 \\
  Qwen3-VL 8B Instruct~\cite{Qwen3-vl}  &28.84 &24.63 &36.95 &0.00 &38.09 &19.23 &39.86 &10.00 &8.10 &38.18 &23.52 &\textbf{42.10} &26.86 &4.54 &30.65 \\
  Qwen3-VL 235B Thinking~\cite{Qwen3-vl}  &22.11 &14.49 &33.33 &0.00 &41.26 &23.07 &38.56 &15.00 &35.13 &38.18 &\textbf{32.35} &\textbf{42.10} &29.26 &40.90 &31.23 \\
  Qwen3-VL 235B Instruct~\cite{Qwen3-vl}  &25.00 &24.63 &39.13 &0.00 &42.85 &26.92 &35.94 &10.00 &5.40 &28.18 &26.47 &\textbf{42.10} &24.39 &9.09 &29.13 \\
  GLM-4.1V-Thinking~\cite{glm}  &0.00 &0.00 &0.00 &0.00 &0.00 &0.00 &0.00 &0.00 &0.00 &0.00 &0.00 &0.00 &0.00 &0.00 &0.00 \\
  GLM-4.5V~\cite{glm}  &0.00 &0.00 &0.74 &0.00 &0.00 &0.00 &0.00 &0.00 &0.00 &0.00 &0.00 &0.00 &0.00 &0.00 &0.12 \\
  \rowcolor{blue!8} \multicolumn{16}{l}{\textit{Embodied Planning Models}} \\
  RoboBrain 2.0-7B~\cite{robobrain2}  &27.27 &\textbf{25.67} &34.31 &0.00 &45.16 &53.33 &34.86 &\textbf{22.72} &5.26 &40.77 &27.77 &36.84 &37.03 &9.09 &31.77 \\
  RoboBrain 2.0-32B~\cite{robobrain2}  &23.14 &20.27 &28.99 &0.00 &22.58 &13.33 &26.31 &18.18 &7.89 &18.44 &19.44 &26.31 &25.92 &9.09 &21.96 \\
  ApBot~\cite{apbot}  &0.80 &2.60 &4.40 &0.00 &18.20 &25.00 &3.70 &2.80 &5.80 &9.90 &16.70 &26.30 &5.90 &0.00 &7.00 \\
  \rowcolor{gray!10}
  \model (Ours)  & \textbf{31.40} & 22.97 & 34.91 & 0.00 & \textbf{54.84} & \textbf{56.67} & \textbf{42.76} & 18.18 & 31.58 & \textbf{63.11} & 22.22 & 31.58 & 25.93 & 4.55 & \textbf{37.12} \\
  \rowcolor{gray!10}
  - \textit{Manual-App. Part Alignment} & 34.71 & 32.43 & 36.09 & 0.00 & 53.23 & 20.00 & 43.42 & 9.09 & 10.53 & 47.57 & 13.89 & 47.37 & 37.04 & 0.00 & 34.67 \\
  \rowcolor{gray!10}
  - \textit{Key-step Action Prediction} & 40.50 & 28.38 & 36.09 & 0.00 & 51.61 & 53.33 & 48.68 & 27.27 & 21.05 & 40.78 & 38.89 & 42.11 & 37.04 & 9.09 & 38.24 \\
  \rowcolor{gray!10}
  - \textit{Part State Judgement} & 33.06 & 27.03 & 33.73 & 0.00 & 56.45 & 23.33 & 42.76 & 13.64 & 13.16 & 48.54 & 8.33 & 47.37 & 37.04 & 4.55 & 34.00 \\
  \bottomrule
        \end{tabular}}
    }
    \label{tab:single_task}
    \end{center}
\end{table*}

%% file: tables/sequential_plan.tex
\begin{table*}[tb]
    \begin{center}
    \small
    \caption{\textbf{Sequential planning and adjustment results based on \benchmark.} This evaluation-only setting combines open-loop planning with online closed-loop correction. Entries report task completion rate / task success rate.}
    \resizebox{\textwidth}{!}{
        \setlength{\tabcolsep}{1.0mm}{
        \begin{tabular}{r|c|c|c|c|c|c|c|c|c|c|c|c|c|c|c}
  \toprule
    \multirow{1}{*}{\centering \textbf{\textsc{Baseline Model}}}
    & \includegraphics[width=0.025\linewidth]{./icon/oven.png}
    & \includegraphics[width=0.025\linewidth]{./icon/toaster.png}
    & \includegraphics[width=0.025\linewidth]{./icon/airfryer.png}
    & \includegraphics[width=0.025\linewidth]{./icon/mixer.png}
    & \includegraphics[width=0.025\linewidth]{./icon/washing.png}
    & \includegraphics[width=0.025\linewidth]{./icon/icemaker.png}
    & \includegraphics[width=0.025\linewidth]{./icon/micro.png}
    & \includegraphics[width=0.025\linewidth]{./icon/ricecooker.png}
    & \includegraphics[width=0.025\linewidth]{./icon/hotpot.png}
    & \includegraphics[width=0.025\linewidth]{./icon/blender.png}
    & \includegraphics[width=0.025\linewidth]{./icon/coffee_machine.png}
    & \includegraphics[width=0.025\linewidth]{./icon/kettle.png}
    & \includegraphics[width=0.025\linewidth]{./icon/breadmachine.png}
    & \includegraphics[width=0.025\linewidth]{./icon/deepfryer.png}
    & \textbf{Total} \\
  \toprule
  \multicolumn{15}{c}{\textbf{Sequential Planning and Adjustment} (\emph{Task Completion Rate/Task Success Rate})} \\
  \midrule
  \rowcolor{blue!8} \textit{Proprietary MLLMs} & & & & & & & & & & & & & & & \\
  GPT-5~\cite{GPT-5}  &2.70/1.06 &4.42/1.13 &1.68/0.00 &0.00/0.00 &7.07/1.69 &4.70/0.00 &4.39/0.73 &0.69/0.00 &4.44/1.85 &2.38/0.00 &5.90/1.85 &5.18/0.00 &14.55/7.01 &14.00/13.33 &4.26/1.35  \\
  GPT-5 Mini~\cite{GPT-5}  &2.22/2.22 &8.02/2.50 &0.11/0.00 &0.00/0.00 &3.58/0.00 &4.44/0.00 &3.46/1.63 &5.86/0.00 &6.10/4.00 &2.31/0.00 &4.99/0.99 &5.55/0.00 &9.29/4.00 &16.99/13.33 &3.98/1.51 \\
  Gemini 2.5 Pro~\cite{gemini}  &0.58/0.00 &4.32/1.12 &3.51/1.45 &0.00/0.00 &2.94/0.00 &4.44/0.00 &5.14/4.68 &0.37/0.00 &0.47/0.00 &1.98/0.80 &11.17/8.41 &5.92/0.00 &16.48/12.96 &14.66/13.33 &4.58/2.81  \\
  Gemini 2.5 Flash~\cite{gemini}  &1.34/1.11 &2.68/0.00 &6.29/3.78 &0.00/0.00 &4.94/2.00 &2.33/0.00 &3.08/2.38 &6.90/2.94 &4.00/3.84 &2.30/0.86 &11.11/8.73 &5.55/0.00 &15.20/11.76 &13.33/13.33 &4.97/3.25  \\
  \rowcolor{blue!8} \textit{Open-source MLLMs} & & & & & & & & & & & & & & & \\
  Qwen3-VL 235B Thinking~\cite{Qwen3-vl}   &2.64/2.19 &9.24/4.70 &1.57/0.00 &0.00/0.00 &9.46/4.00 &7.83/3.70 &2.82/2.41 &8.99/2.77 &7.25/3.92 &4.03/0.85 &6.61/4.00 &8.33/0.00 &8.97/3.77 &11.33/10.00 &5.05/2.48 \\
  Qwen3-VL 235B Instruct~\cite{Qwen3-vl}  &3.33/3.33 &6.86/1.25 &4.00/3.84 &0.00/0.00 &10.00/6.00 &6.03/3.70 &5.28/4.95 &3.14/0.00 &5.13/4.00 &2.47/1.76 &9.62/6.06 &8.88/0.00 &19.95/16.00 &10.00/10.00 &5.84/4.08    \\
  Qwen3-VL 8B Thinking~\cite{Qwen3-vl}   &1.55/1.11 &6.23/2.50 &1.51/1.51 &0.00/0.00 &8.17/6.00 &4.76/3.70 &1.84/1.65 &7.42/3.33 &8.36/3.92 &0.88/0.88 &7.07/5.05 &2.77/0.00 &11.22/8.00 &10.00/10.00 &4.03/2.74  \\
  Qwen3-VL 8B Instruct~\cite{Qwen3-vl}  &0.00/0.00 &7.26/0.00 &3.07/3.07 &0.00/0.00 &2.00/2.00 &0.00/0.00 &1.66/1.66 &0.00/0.00 &4.00/4.00 &0.88/0.88 &2.69/1.01 &12.22/0.00 &6.71/4.65 &6.66/6.66 &2.65/1.54  \\
  GLM-4.1V-Thinking~\cite{glm}  &0.00/0.00 &0.00/0.00 &0.00/0.00 &0.00/0.00 &0.00/0.00 &0.00/0.00 &0.00/0.00 &0.00/0.00 &0.00/0.00 &0.00/0.00 &0.00/0.00 &0.00/0.00 &0.00/0.00 &0.00/0.00 &0.00/0.00    \\
  GLM-4.5V~\cite{glm}   &2.22/2.22 &3.30/1.25 &2.39/2.29 &1.11/1.11 &5.21/4.00 &4.44/3.70 &1.30/0.83 &1.33/0.00 &4.43/2.00 &1.50/0.97 &8.33/6.00 &3.33/0.00 &10.11/8.00 &12.66/10.00 &3.72/2.67    \\
  \rowcolor{blue!8} \textit{Embodied Planning Models} & & & & & & & & & & & & & & & \\
  RoboBrain 2.0-7B~\cite{robobrain2}  &0.00/0.00 &0.00/0.00 &0.06/0.00 &0.00/0.00 &0.00/0.00 &0.00/0.00 &0.16/0.00 &0.00/0.00 &0.00/0.00 &0.27/0.00 &0.00/0.00 &0.52/0.00 &0.20/0.00 &0.00/0.00 &0.08/0.00    \\
  RoboBrain 2.0-32B~\cite{robobrain2}  &0.00/0.00 &0.41/0.00 &0.00/0.00 &0.00/0.00 &2.43/2.43 &0.00/0.00 &0.00/0.00 &0.00/0.00 &0.00/0.00 &0.00/0.00 &1.96/1.96 &0.00/0.00 &0.00/0.00 &0.00/0.00 &0.37/0.30    \\
  ApBot~\cite{apbot}  &0.00/0.00 &0.00/0.00 &0.00/0.00 &0.00/0.00 &2.000/2.00 &9.20/7.40 &1.30/0.80 &0.00/0.00 &0.00/0.00 &5.30/5.30 &5.10/5.10 &0.00/0.00 &2.40/2.00 &0.00/0.00 &1.80/1.60    \\
  \rowcolor{gray!10}
  \model (Ours)  & \textbf{43.94/24.44} & \textbf{35.23/21.25} & \textbf{48.19/36.22} & \textbf{46.83/16.67} & \textbf{29.46/14.00} & \textbf{44.44/44.44}& \textbf{40.96/24.17} & \textbf{17.92/13.33} & \textbf{32.71/10.00} & \textbf{45.29/30.09} & \textbf{59.45/46.46} & \textbf{65.00/50.00} & \textbf{55.65/36.00} & \textbf{52.17/30.00} & \textbf{44.59/28.07} \\
  \rowcolor{gray!10}
  - \textit{Manual-App. Part Alignment} & 43.49/25.56 & 31.31/16.25 & 43.87/33.07 & 47.14/22.22 & 34.72/16.00 & 46.46/25.93 & 31.71/12.50 & 23.75/6.67 & 37.21/18.00 & 48.96/35.40 & 51.44/39.39 & 75.00/65.00 & 61.31/34.00 & 53.75/36.67 & 43.45/26.54 \\
  \rowcolor{gray!10}
  - \textit{Key-step Action Prediction} & 34.65/14.44 & 30.68/16.25 & 30.48/19.69 & 45.48/17.78 & 20.13/12.00 & 27.41/14.81 & 31.86/14.17 & 21.27/6.67 & 36.29/10.00 & 49.69/30.97 & 46.20/20.20 & 61.67/45.00 & 64.84/36.00 & 54.58/40.00 & 38.79/19.98 \\
  \rowcolor{gray!10}
  - \textit{Part State Judgement} & 43.99/23.33 & 30.71/15.00 & 32.91/25.20 & 52.38/30.00 & 31.73/16.00 & 36.40/22.22 & 35.49/15.83 & 8.75/6.67 & 23.38/14.00 & 48.37/33.63 & 51.53/35.35 & 70.00/60.00 & 53.62/30.00 & 51.25/36.67 & 40.73/25.10 \\
  \bottomrule
        \end{tabular}}
    }
    \label{tab:sequential}
    \end{center}
\end{table*}

%% file: tables/real_robot.tex
\begin{table}[H]
\scriptsize
\centering
\setlength{\tabcolsep}{2pt}
\caption{\textbf{Real-robot results on household appliances.} Entries report task completion rate/task success rate.}
\label{tab:real_robot}
\resizebox{\columnwidth}{!}{
\begin{tabular}{l l c c c c c c}
\toprule
& & {\scriptsize\texttt{Microwave}} & {\scriptsize\texttt{Air-fryer}} & {\scriptsize\texttt{Ricecooker}} & {\scriptsize\texttt{Coffee Machine}} & {\scriptsize\texttt{Toaster}} & {\scriptsize\texttt{Blender}} \\
\midrule
GPT-5 & & 16.67/0.00 & 11.33/0.00 & 28.30/10.00 & 18.30/0.00 & 14.90/0.00 & 26.60/10.00 \\
ApBot~\citep{apbot} & & 11.80/0.00 & 14.70/0.00 & 3.30/0.00 & 8.30/10.00 & 3.30/0.00 & 13.30/10.00 \\
\rowcolor{gray!15}
\textbf{\model (Ours)} & & 50.00/40.00 & 49.10/40.00 & 26.80/20.00 & 52.50/40.00 & 33.33/20.00 & 87.50/80.00 \\
\bottomrule
\end{tabular}
}
\end{table}

%% file: reference.bib
@misc{realappliance,
      title={RealAppliance: Let High-fidelity Appliance Assets Controllable and Workable as Aligned Real Manuals},
      author={Yuzheng Gao and Yuxing Long and Lei Kang and Yuchong Guo and Ziyan Yu and Shangqing Mao and Jiyao Zhang and Ruihai Wu and Dongjiang Li and Hui Shen and Hao Dong},
      year={2026},
      archivePrefix={arXiv},
      primaryClass={cs.CV}
}

@inproceedings{clip,
  title={Learning Transferable Visual Models From Natural Language Supervision},
  author={Radford, Alec and Kim, Jong Wook and Hallacy, Chris and Ramesh, Aditya and Goh, Gabriel and Agarwal, Sandhini and Sastry, Girish and Askell, Amanda and Mishkin, Pamela and Clark, Jack and Krueger, Gretchen and Sutskever, Ilya},
  booktitle={Proceedings of the 38th International Conference on Machine Learning},
  pages={8748--8763},
  year={2021}
}

@inproceedings{deepspeed,
  title={DeepSpeed: System Optimizations Enable Training Deep Learning Models with Over 100 Billion Parameters},
  author={Rasley, Jeff and Rajbhandari, Samyam and Ruwase, Olatunji and He, Yuxiong},
  booktitle={Proceedings of the 26th ACM SIGKDD International Conference on Knowledge Discovery \& Data Mining},
  pages={3505--3506},
  year={2020}
}

@article{anygrasp,
  title={Anygrasp: Robust and efficient grasp perception in spatial and temporal domains},
  author={Fang, Hao-Shu and Wang, Chenxi and Fang, Hongjie and Gou, Minghao and Liu, Jirong and Yan, Hengxu and Liu, Wenhai and Xie, Yichen and Lu, Cewu},
  journal={IEEE Transactions on Robotics},
  volume={39},
  number={5},
  pages={3929--3945},
  year={2023},
  publisher={IEEE}
}

@inproceedings{foundationpose,
  title={Foundationpose: Unified 6d pose estimation and tracking of novel objects},
  author={Wen, Bowen and Yang, Wei and Kautz, Jan and Birchfield, Stan},
  booktitle={Proceedings of the IEEE/CVF Conference on Computer Vision and Pattern Recognition},
  pages={17868--17879},
  year={2024}
}

@misc{glm,
      title={GLM-4.5V and GLM-4.1V-Thinking: Towards Versatile Multimodal Reasoning with Scalable Reinforcement Learning}, 
      author={GLM-V Team},
      year={2025},
      eprint={2507.01006},
      archivePrefix={arXiv},
      primaryClass={cs.CV},
      url={https://arxiv.org/abs/2507.01006}, 
}

@article{qwen25vl,
  title={Qwen2. 5-vl technical report},
  author={Bai, Shuai and Chen, Keqin and Liu, Xuejing and Wang, Jialin and Ge, Wenbin and Song, Sibo and Dang, Kai and Wang, Peng and Wang, Shijie and Tang, Jun and others},
  journal={arXiv preprint arXiv:2502.13923},
  year={2025}
}

@inproceedings{checkmanual,
  author    = {Long, Yuxing and Zhang, Jiyao and Pan, Mingjie and Wu, Tianshu and Kim, Taewhan and Dong, Hao},
  title     = {CheckManual: A New Challenge and Benchmark for Manual-based Appliance Manipulation},
  booktitle = {Proceedings of the IEEE/CVF Conference on Computer Vision and Pattern Recognition (CVPR)},
  month     = {June},
  year      = {2025}
}

@article{qwenimage,
  title={Qwen-image technical report},
  author={Wu, Chenfei and Li, Jiahao and Zhou, Jingren and Lin, Junyang and Gao, Kaiyuan and Yan, Kun and Yin, Sheng-ming and Bai, Shuai and Xu, Xiao and Chen, Yilei and others},
  journal={arXiv preprint arXiv:2508.02324},
  year={2025}
}

@article{seedream4,
  title={Seedream 4.0: Toward next-generation multimodal image generation},
  author={Seedream, Team and Chen, Yunpeng and Gao, Yu and Gong, Lixue and Guo, Meng and Guo, Qiushan and Guo, Zhiyao and Hou, Xiaoxia and Huang, Weilin and Huang, Yixuan and others},
  journal={arXiv preprint arXiv:2509.20427},
  year={2025}
}

@article{liu2023grounding,
  title={Grounding dino: Marrying dino with grounded pre-training for open-set object detection},
  author={Liu, Shilong and Zeng, Zhaoyang and Ren, Tianhe and Li, Feng and Zhang, Hao and Yang, Jie and Li, Chunyuan and Yang, Jianwei and Su, Hang and Zhu, Jun and others},
  journal={arXiv preprint arXiv:2303.05499},
  year={2023}
}

@misc{openxe,
title={Open {X-E}mbodiment: Robotic Learning Datasets and {RT-X} Models},
author = {Open X-Embodiment Collaboration},
howpublished  = {\url{https://arxiv.org/abs/2310.08864}},
year = {2023},
}

@article{apbot,
  title={Robot Operation of Home Appliances by Reading User Manuals},
  author={Zhang, Jian and Zhang, Hanbo and Xiao, Anxing and Hsu, David},
  journal={arXiv preprint arXiv:2505.20424},
  year={2025}
}

@article{black2024pi_0,
  title={pi0: A Vision-Language-Action Flow Model for General Robot Control},
  author={Black, Kevin and Brown, Noah and Driess, Danny and Esmail, Adnan and Equi, Michael and Finn, Chelsea and Fusai, Niccolo and Groom, Lachy and Hausman, Karol and Ichter, Brian and others},
  journal={arXiv preprint arXiv:2410.24164},
  year={2024}
}

@inproceedings{li2024manipllm,
  title={Manipllm: Embodied multimodal large language model for object-centric robotic manipulation},
  author={Li, Xiaoqi and Zhang, Mingxu and Geng, Yiran and Geng, Haoran and Long, Yuxing and Shen, Yan and Zhang, Renrui and Liu, Jiaming and Dong, Hao},
  booktitle={Proceedings of the IEEE/CVF Conference on Computer Vision and Pattern Recognition},
  pages={18061--18070},
  year={2024}
}

@article{robobrain2,
  title={Robobrain 2.0 technical report},
  author={Team, BAAI RoboBrain and Cao, Mingyu and Tan, Huajie and Ji, Yuheng and Chen, Xiansheng and Lin, Minglan and Li, Zhiyu and Cao, Zhou and Wang, Pengwei and Zhou, Enshen and others},
  journal={arXiv preprint arXiv:2507.02029},
  year={2025}
}

@inproceedings{robobrain,
  title={Robobrain: A unified brain model for robotic manipulation from abstract to concrete},
  author={Ji, Yuheng and Tan, Huajie and Shi, Jiayu and Hao, Xiaoshuai and Zhang, Yuan and Zhang, Hengyuan and Wang, Pengwei and Zhao, Mengdi and Mu, Yao and An, Pengju and others},
  booktitle={Proceedings of the Computer Vision and Pattern Recognition Conference},
  pages={1724--1734},
  year={2025}
}

@article{GPT-4,
  title={Gpt-4 technical report},
  author={OpenAI},
  journal={arXiv:2303.08774},
  year={2023}
}

@inproceedings{llava,
  title={Visual instruction tuning},
  author={Liu, Haotian and Li, Chunyuan and Wu, Qingyang and Lee, Yong Jae},
  booktitle={NeurIPS},
  year={2023}
}

@inproceedings{blip2,
  title={BLIP-2: Bootstrapping language-image pre-training with frozen image encoders and large language models},
  author={Li, Junnan and Li, Dongxu and Savarese, Silvio and Hoi, Steven},
  booktitle={ICML},
  year={2023}
}

@article{flamingo,
  title={Flamingo: a visual language model for few-shot learning},
  author={Alayrac, Jean-Baptiste and Donahue, Jeff and Luc, Pauline and Miech, Antoine and Barr, Iain and Hasson, Yana and Lenc, Karel and Mensch, Arthur and Millican, Katherine and Reynolds, Malcolm and others},
  journal={NeurIPS},
  year={2022}
}

@misc{gemini,
  author       = {{Google AI}},
  title        = {Gemini 2.5 API},
  year         = 2025,
  url          = {https://ai.google.dev/gemini-api/docs}
}

@misc{GPT-5,
  title  ={Gpt5},
  author={OpenAI},
  year         = 2025,
  url          = {https://openai.com/index/introducing-gpt-5/}
}

@misc{Qwen3-vl,
    title        = {Qwen3-VL API},
    author       = {{Qwen Team}},
    year         = {2025},
    url          = {https://modelstudio.console.alibabacloud.com/?tab=doc#/doc/?type=model&url=2840914_2&modelId=qwen3-vl-plus}
}

@misc{OmniManip,
      title={OmniManip: Towards General Robotic Manipulation via Object-Centric Interaction Primitives as Spatial Constraints}, 
      author={Mingjie Pan and Jiyao Zhang and Tianshu Wu and Yinghao Zhao and Wenlong Gao and Hao Dong},
      year={2025},
      eprint={2501.03841},
      archivePrefix={arXiv},
      primaryClass={cs.RO},
      url={https://arxiv.org/abs/2501.03841}, 
}

@article{saycan,
  title={Do As I Can, Not As I Say: Grounding Language in Robotic Affordances},
  author={Ahn, Michael and Brohan, Anthony and Brown, Noah and Chebotar, Yevgen and Cortes, Ricardo and David, Byron and Finn, Chelsea and Fu, Chuyuan and Gopalakrishnan, Keerthana and Hausman, Karol and others},
  journal={arXiv preprint arXiv:2204.01691},
  year={2022}
}

@article{palme,
  title={PaLM-E: An Embodied Multimodal Language Model},
  author={Driess, Danny and Xia, Fei and Sajjadi, Mehdi S. M. and Lynch, Corey and Chowdhery, Aakanksha and Ichter, Brian and Wahid, Ayzaan and Tompson, Jonathan and Vuong, Quan and Yu, Tianhe and others},
  journal={arXiv preprint arXiv:2303.03378},
  year={2023}
}

@article{rt2,
  title={RT-2: Vision-Language-Action Models Transfer Web Knowledge to Robotic Control},
  author={Brohan, Anthony and Brown, Noah and Carbajal, Justice and Chebotar, Yevgen and Driess, Danny and Finn, Chelsea and Fu, Chuyuan and Gopalakrishnan, Keerthana and Hausman, Karol and Herzog, Alex and others},
  journal={arXiv preprint arXiv:2307.15818},
  year={2023}
}

@article{openvla,
  title={OpenVLA: An Open-Source Vision-Language-Action Model},
  author={Kim, Moo Jin and Pertsch, Karl and Karamcheti, Siddharth and Xiao, Ted and Balakrishna, Ashwin and Nair, Suraj and Rafailov, Rafael and Foster, Ethan and Lam, Grace and Sanketi, Pannag and others},
  journal={arXiv preprint arXiv:2406.09246},
  year={2024}
}

@inproceedings{react,
  title={ReAct: Synergizing Reasoning and Acting in Language Models},
  author={Yao, Shunyu and Zhao, Jeffrey and Yu, Dian and Du, Nan and Shafran, Izhak and Narasimhan, Karthik and Cao, Yuan},
  booktitle={International Conference on Learning Representations (ICLR)},
  year={2023}
}
